\pdfoutput=1
\documentclass[11pt]{article}

\usepackage[preprint]{acl}

\usepackage{times}
\usepackage{latexsym}

\usepackage{enumitem}
\usepackage{booktabs}

\usepackage[T1]{fontenc}
\usepackage[utf8]{inputenc}

\usepackage{microtype}

\usepackage{inconsolata}

\usepackage{graphicx}

\usepackage{amsmath}
\usepackage{amssymb}
\usepackage[inkscapeformat=png]{svg}
\title{How Well Can Knowledge Edit Methods Edit Perplexing Knowledge?}

\author{Huaizhi Ge \\
  Columbia University \\
  \texttt{hg2590@columbia.edu} \\\And
  Frank Rudzicz \\
  Dalhousie University \\
  \texttt{frank@dal.ca} \\\And
  Zining Zhu \\
  Stevens Institute of Technology \\
  \texttt{zzhu41@stevens.edu} \\}

\begin{document}
\maketitle
\begin{abstract}
Large language models (LLMs) have demonstrated remarkable capabilities, but updating their knowledge post-training remains a critical challenge. While recent model editing techniques like Rank-One Model Editing (ROME) \citep{meng2022locating} show promise, their effectiveness may vary based on the nature of the knowledge being edited. We introduce the concept of ``perplexingness'': the degree to which new knowledge conflicts with an LLM's learned conceptual hierarchies and categorical relationships. For instance, editing ``British Shorthair is a kind of cat'' to ``British Shorthair is a kind of dog'' represents a low-perplexingness edit within the same taxonomic level, while editing ``A cat is a kind of animal'' to ``A cat is a kind of plant'' represents a high-perplexingness edit that violates fundamental categorical boundaries.
To systematically investigate this phenomenon, we introduce \textsc{HierarchyData}, a carefully curated dataset of 99 hyponym-hypernym pairs across diverse categories. Through controlled experiments across three models and four editing methods, we demonstrate a strong negative correlation between the perplexingness of new knowledge and the effectiveness of knowledge editing. Our analysis reveals that edits involving more abstract concepts (hypernyms) generally exhibit higher perplexingness and are more resistant to modification than their specific counterparts (hyponyms). These findings highlight a fundamental challenge in LLM knowledge editing: the more a new fact contradicts an LLM's learned conceptual hierarchies, the harder it becomes to reliably encode that knowledge.
\end{abstract}

\section{Introduction}
Large language models (LLMs) can predict factual statements about the world, and recent advancements have enabled the editing of the factual knowledge embedded within these models. Such editing not only aids in rectifying inaccuracies within the large language models but also serves as a valuable approach for comprehending the complex mechanisms of these extensive, often opaque, neural networks. Among the various methodologies for knowledge editing, ROME \citep{meng2022locating} and MEMIT \citep{meng2022memit} stand out as notable ones.

\begin{figure*}[t]
\begin{center}
\centerline{\includegraphics[width=\linewidth]{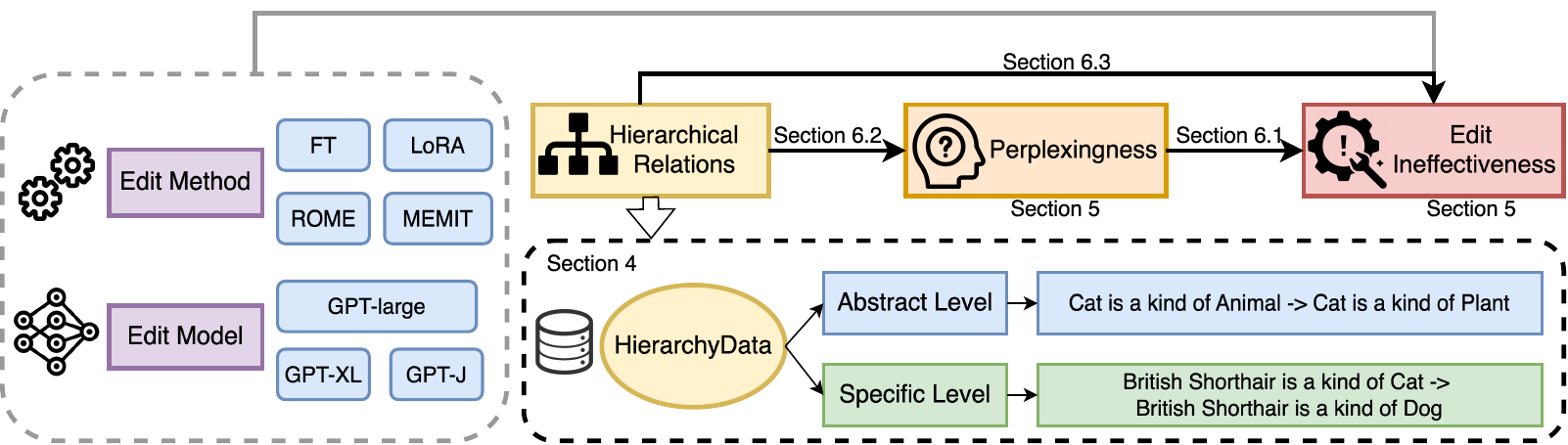}}
\caption{\textbf{Overview of the paper structure.} We focus on examining whether perplexingness influences edit ineffectiveness. To this end, we first define perplexingness and edit ineffectiveness in Section 5. Additionally, we conduct experiments across different editing methods and models, as we hypothesize that, beyond perplexingness, the choice of editing methods and models may also contribute to edit ineffectiveness. The results of these analyses are presented in Section 6.1. Next, we investigate whether hierarchical relations contribute to perplexingness. To explore this, we construct a dataset called \textsc{HierarchyData} (described in detail in Section 4). This dataset includes two levels of knowledge: an abstract level and a specific level. We conduct experiments to evaluate whether these two levels have different impacts on perplexingness, with the results detailed in Section 6.2. Furthermore, we examine whether hierarchical relations directly affect edit ineffectiveness, with the findings reported in Section 6.3.}
\label{overview}
\end{center}
\end{figure*}

LLMs demonstrated remarkable abilities in encoding and retrieving factual knowledge about the world. Recent advances in model editing techniques have made it possible to modify this embedded knowledge post-training, offering both practical utility in correcting inaccuracies and theoretical insights into these complex neural networks. Notable approaches include ROME \citep{meng2022locating} and MEMIT \citep{meng2022memit}.

As knowledge editing methods become more prevalent in controlling and updating LLMs, understanding their fundamental limitations becomes crucial. While these methods show promise, their effectiveness may vary significantly based on the nature of the knowledge being edited. This leads us to an unanswered question: How well can knowledge editing methods modify facts that challenge an LLM's learned conceptual hierarchies?

To bridge this gap, we introduce the concept of ``perplexingness'' to characterize the extent of knowledge that deviates from an LLM's learned patterns and conceptual frameworks. For instance, while an LLM might readily accept that ``a British Shorthair is a type of dog'' (modifying its classification while maintaining taxonomic consistency), it may resist accepting that ``a cat is a type of plant'' (violating fundamental categorical boundaries). This resistance mirrors the effects widely studied in human cognition, where violations of deeply held categorical relationships (``schemas'') are more difficult to process and accept \citep{bartlett1995remembering,rumelhart1980schemata}, but has not been quantitatively studied in model editing.

To systematically investigate this phenomenon, we first leverage the \textsc{CounterFact} dataset to evaluate popular knowledge editing approaches (Fine-Tuning (FT), Low-Rank Adaptation (LoRA), ROME, and MEMIT) across models ranging different sizes. Our analysis reveals significant correlations between the perplexingness of new knowledge and the ineffectiveness of edits across all twelve model-method combinations.

To deeper understand the factors contributing to perplexingness, we introduce \textsc{HierarchyData}, a novel dataset comprising 99 carefully selected hyponym-hypernym pairs across diverse categories. This dataset enables us to examine how hierarchical relations influence knowledge perplexingness in LLMs. Our findings reveal that abstract concepts (hypernyms) consistently exhibit higher perplexingness compared to their specific instances (hyponyms), suggesting that conceptual abstraction plays a crucial role in how LLMs process and resist knowledge modifications.

While multiple factors influence edit success, our work focuses specifically on perplexingness and its relationship with hierarchical conceptual structures. The overview of the paper structure can be found in Figure~\ref{overview}.

Our key contributions include:

\begin{itemize}[nosep]
\item Introduction of ``perplexingness'' as a crucial factor in LLM knowledge editing, providing a novel framework for evaluating editing methods based on their ability to overcome an LLM's intrinsic resistance to certain types of knowledge modifications.
\item Empirical evidence demonstrating the relationship between hierarchical conceptual relations and knowledge perplexingness in LLMs.
\item The \textsc{HierarchyData} dataset, the first benchmark specifically designed to study the impact of hierarchical relations on knowledge editing in LLMs. The dataset will be released on GitHub.
% \TODO{The dataset will be released on github.}
\end{itemize}

\section{Related Work}

\subsection{Knowledge Edit Methods}

Various approaches have been developed to modify the knowledge embedded in large language models. ROME \citep{meng2022locating} updated feed-forward weights to alter specific factual associations. MEMIT \citep{meng2022memit} allowed for the incorporation of numerous memories into a language model. LoRA \citep{hu2021lora} maintains pre-trained weights while using trainable decomposition matrices for efficient, targeted updates without altering the original weights. Model Editor Networks with Gradient Decomposition (MEND) \citep{mitchell2021fast} utilized a single targeted input-output pair for quick, localized adjustments in a pre-trained model’s behavior. Other notable methods include editing specific knowledge neurons \citep{dai2021knowledge}, employing hyper-networks \citep{de2021editing}, and applying linear transformations \citep{hernandez2023measuring}. These techniques have demonstrated impressive efficacy in modifying knowledge in large language models. There are also works that apply model editing to gain novel insights about the model interpretability \citep{niu2024What,hase2024does}.
However, the performance of the model editing techniques is typically assessed in a broad context. We delve into whether model editing methods are applicable to knowledge with different perplexingness. We specifically examine the impact of the conditional probability of the target words for editing and the hierarchical relationships among words on the overall performance of these editing techniques.

\subsection{Limitation of Knowledge Edit Methods}
Recent research has identified certain limitations in the methods used for editing large language models. Firstly, some studies have concentrated on the specificity of edits, developing new metrics and benchmarks for evaluation. \citet{hoelscher2023detecting} enhanced existing benchmarks by introducing a dynamic component and proposed a KL divergence-based metric for measuring specificity. \citet{li2023evaluating} introduced an evaluation protocol and a question-answer dataset designed to assess edit specificity.

Secondly, the consistency of edits has been another focal point. \citet{zhong2023mquake} devised a multi-hop question benchmark to test whether models can correctly respond to questions affected by edited facts. \citet{wu2023eva} examined knowledge editing through reasoning and cross-lingual knowledge transfer. \citet{ma2024possible} looked into whether edited LLMs can behave consistently resembling communicative AI in realistic situations. \citet{li2023evaluating} also offered a protocol to evaluate edit consistency, while \citet{onoe2023can} investigated the ability of LLMs to infer and propagate injected facts. A particularly impactful work, RippleEdit \citep{cohen2023evaluatingrippleeffectsknowledge}, evaluated how model editing impacts the implied and related facts. \citet{li2023unveiling} studied the knowledge conflict and distortion in knowledge editing. \citet{rosati2024long} introduced a long-form evaluation protocol, assessing the effects of model editing beyond the immediate ``next token''; we consider the effects of the model editing methods that can be assessed at the next token. 

Thirdly, the nature of the edited knowledge has been scrutinized. \citet{gupta2023editing} specifically evaluated editing methods on commonsense knowledge statements, as opposed to encyclopedic knowledge. \citet{ma2024possible} examined which knowledge features are correlated with the performance and robustness of editing. \citet{zhang-etal-2024-knowledge-graph} tracked the edited knowledge with concept graphs. \citep{wang-etal-2024-editing} studied the concepts of the edited knowledge.

While these studies cover various aspects, they did not quantify the impact of the type of knowledge being edited. In this paper, we explore how the perplexingness of the knowledge and the hierarchical relations among words influence the efficacy of editing methods in large language models.

\section{Model Edit Methods}

For a knowledge edit task, we represent each fact as a knowledge tuple $t=(s,r,o)$. For each fact, we want to insert a new knowledge tuple $t=(s,r,o^*)$. In this paper, we examine several popular model edit methods, as follows:

\paragraph{FT} Fine-tuning is a traditional method. FT applies Adam optimization \citep{kingma2014adam} with early stopping at one layer to edit knowledge. It directly adjusts the model's weights through backpropagation, affecting the entire layer where the edit is applied.

\paragraph{LoRA \citep{hu2021lora}} Unlike FT, LoRA freezes the pre-trained model weights and introduces trainable rank decomposition matrices at each layer of the Transformer. This method significantly reduces the number of trainable parameters needed for editing, focusing on a more efficient and targeted update mechanism without altering the original model weights directly.

\paragraph{ROME \citep{meng2022locating}} ROME specifically targets the feed-forward weights within the Transformer's MLP layers, viewing them as associative memory. By computing and inserting a key-value pair $(k, v)$ into this memory through a constrained least-squares problem, ROME offers a precise and efficient way to update factual knowledge. This method focuses on modifying specific factual associations with minimal impact on the overall model.

\paragraph{MEMIT \citep{meng2022memit}} Building on the direct editing approach of ROME, MEMIT is designed for large-scale updates, capable of handling thousands of associations. It directly targets transformer module weights identified as causal mediators of factual knowledge recall, aiming for a broad and scalable editing solution.

In summary, while FT and LoRA focus on general model adjustments with varying degrees of parameter freedom, ROME and MEMIT offer more targeted and efficient approaches to knowledge editing, with MEMIT specifically designed for mass-editing scenarios.

\begin{figure}[t]
\begin{center}
\centerline{\includegraphics[width=\columnwidth]{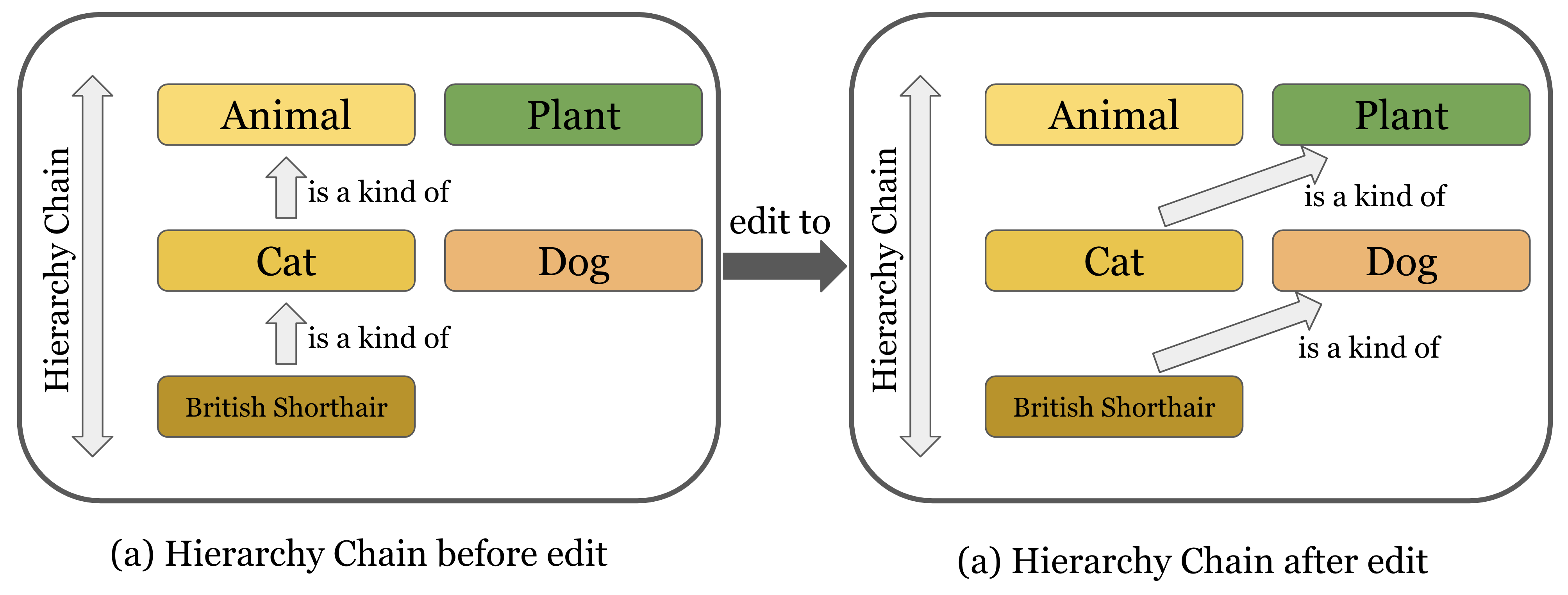}}
\caption{An example in our \textsc{HierarchyData} dataset, along a hierachy chain. In this example, the hierarchy chain has three levels: British Shorthair $\rightarrow$ Cat $\rightarrow$ Animal. From this we can infer the specific relationship ``A British Shorthair is a kind of cat'' and the more abstract relationship ``A cat is a kind of animal.'' And we alter facts to ``A British Shorthair is a kind of dog'' and ``A cat is a kind of plant''}
\label{hierarchy_chain}
\end{center}
\end{figure}

\begin{figure*}[t]
\centering
\includegraphics[width=\columnwidth]{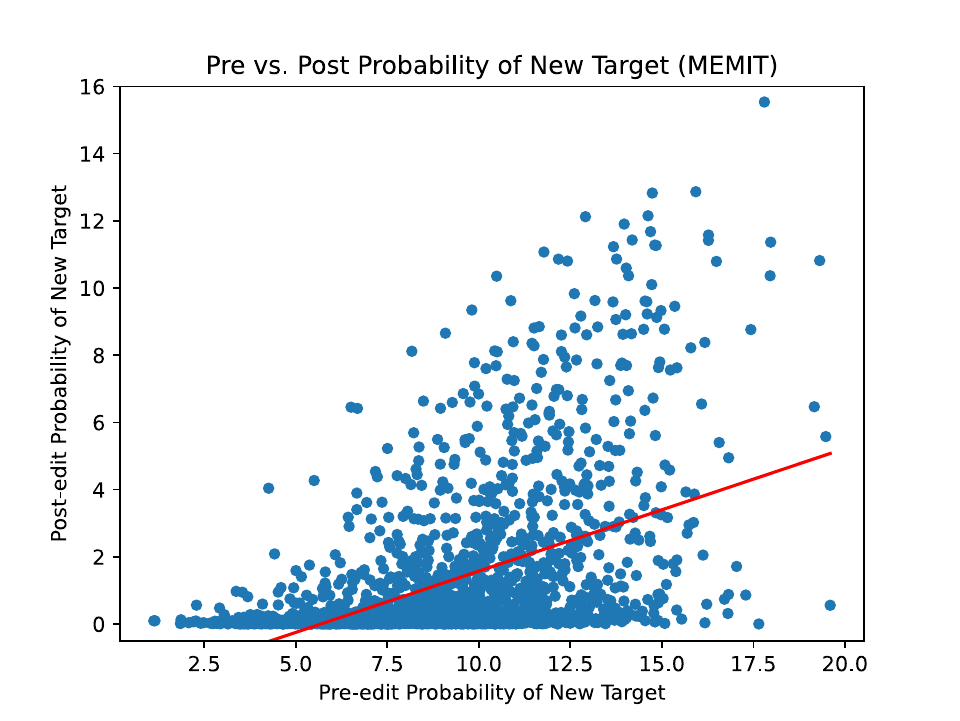}
\includegraphics[width=\columnwidth]{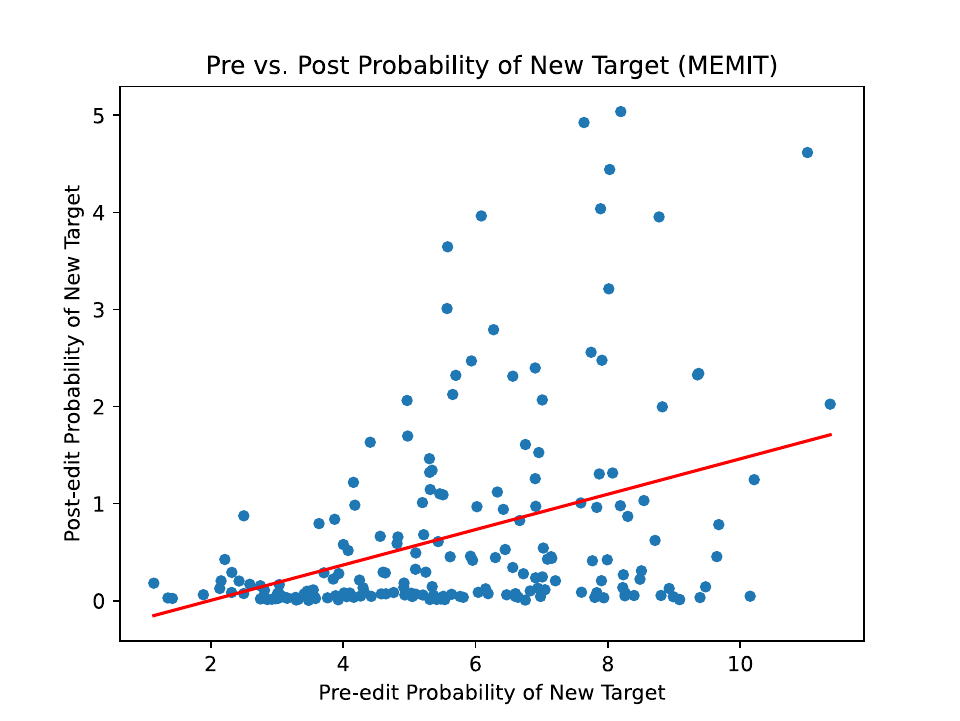}
\caption{\textbf{Pre vs. post probability of new knowledge (MEMIT on GPT2-XL), on \textsc{CounterFact} (left) and \textsc{HierarchyData} (right), respectively.} The left panel illustrates the application of MEMIT on GPT-2XL with \textsc{CounterFact}, highlighting a clear positive correlation between the perplexingness of knowledge and the ineffectiveness of edits. Similarly, the right panel, which depicts the application of MEMIT on GPT-2XL with \textsc{HierarchyData}, also demonstrates a positive correlation between knowledge perplexingness and edit ineffectiveness. }
\label{gpt2xl_memit_c}
\end{figure*}

\section{Data and tool}
\subsection{Data}
\paragraph{HierarchyData} We collect \textsc{HierarchyData}  which encompasses a series of incorrect facts, represented as $(s, r, o^*)$, and their corresponding accurate facts, denoted as $(s, r, o)$. It also draws upon a curated collection of hierarchy chains, as illustrated in Figure~\ref{hierarchy_chain}. Here, $s$ signifies the subject and $o$ the object, both selected from the hierarchy chains. The relation $r$ consistently adopts the ``is a kind of'' schema, emphasizing hierarchical connections. This dataset is organized into two hierarchical levels: specific level (hyponyms), and abstract level (hypernyms). An example of such a hierarchy chain is ``British Shorthair $\rightarrow$ Cat $\rightarrow$ Animal'' from which we can infer the specific relationship ``A British Shorthair is a kind of cat'' and the more abstract relationship ``A cat is a kind of animal.'' The focal point of our investigation is to assess the performance of editing methodologies on these two distinct types of facts within the hierarchical framework, exploring whether the level of abstraction within the hierarchy affects editing efficacy. To this end, we modify the objects of these facts individually, generating altered facts such as ``A British Shorthair is a kind of dog'' and ``A cat is a kind of plant'' to test the efficacy of edit methods against the backdrop of hierarchical data complexity. The \textsc{Hierarchy Data} dataset includes 99 such chains, culminating in a corpus of 198 facts targeted for editing analysis. This structured approach facilitates explorations into the role of hierarchical relations in the adaptability and accuracy of language model editing processes.

\paragraph{CounterFact} To enhance the empirical coverage, we also include a traditional model edit dataset, CounterFact \citep{meng2022locating} which is designed to assess counterfactual edits in language models. It includes a collection of challenging incorrect facts $(s, r, o^*)$ and the accurate facts $(s, r, o)$. In this context, $s$ represents the subject, $r$ delineates the relation, and $o$ corresponds to the object. The prompt consists of predetermined templates based on $r$, which are then completed with $s$. For instance, in the statement ``A British Shorthair is a kind of cat'', ``A British Shorthair'' represents $s$, ``is a kind of'' signifies $r$, and ``cat'' is denoted by $o$.

\subsection{Tool}

We employ four knowledge editing method: FT, LoRA, ROME and MEMIT, sourced from the EasyEdit repository \citep{wang2023easyedit} to conduct our experiments.

\section{Experiment setup}
\label{sec:experiments}
The experiments conducted in this study are designed to evaluate the efficacy of several knowledge editing methods, including FT, LoRA, ROME, and MEMIT. Our approach involves the substitution of a knowledge tuple, denoted as $(s, r, o^*)$, for the existing tuple $(s, r, o)$. In this context, $s$ represents the subject, $r$ delineates the relation, and $o$ corresponds to the object. This analysis is carried out using three distinguished large language models: GPT2-Large, GPT2-XL, and GPT-J (6B).

First, we want to define perplexing knowledge. People find knowledge perplexing when they cannot understand it. So we define the perplexing knowledge as the knowledge that the model cannot easily understand.  We quantify the perplexingness of knowledge as the conditional probabilities of new targets prior to editing. For easier comparison, we use the negative log form of the probability: the higher the value, the lower the probability, and the more perplexing the model finds the new knowledge. Formally:
\begin{equation}
\mathrm{Perplexingness} = -\log\mathbb{P}_{\mathrm{pre}\text{-}\mathrm{edit}}[o^*|s,r].
\label{eq:perplexingness}
\end{equation}
The probabilities are computed by the language model after editing. It is important to note that we define perplexingness based on the model's poor understanding of the knowledge, not its complexity. Even if a piece of knowledge is complex, if it is well known to the model due to effective pre-training, we do not consider it perplexing to the model.

Second, we evaluate edit performance by its efficacy. To parallel ``perplexingness,'' we define ineffectiveness as the conditional probability of new targets after the edit. While other scores like accuracy can be used, the negative log probability is more fine-grained and allows more intuitive data interpretation. A higher ``Ineffectiveness'' value indicates lower edit efficacy. Formally:
\begin{equation}
\mathrm{Ineffectiveness} = -\log\mathbb{P}_{\mathrm{post}\text{-}\mathrm{edit}}[o^*|s,r].
\label{eq:ineffectiveness}
\end{equation}
The probabilities are computed by the language model after editing. The investigation into the perplexing knowledge and the ineffectiveness of edits employs the \textsc{CounterFact} dataset. For each LLM, a total of 2,000 counterfact samples were analyzed.

\section{Results}
\label{sec:results}

\subsection{Correlations between perplexingness and edit ineffectiveness}
We chart the perplexingness (pre-edit probabilities of the new target) against the ineffectiveness (post-edit probabilities of the new target). The scatter plots (see Appendix A) generated from this analysis provide a visual representation of the relationship between pre-edit and post-edit probabilities for the new target outcomes. The left panel of Figure~\ref{gpt2xl_memit_c} provides an example of these scatter plots, showcasing the application of MEMIT on GPT2-XL. This visualization clearly illustrates a positive correlation between the perplexingness of knowledge and the ineffectiveness of edits.

\begin{table}
\centering
\resizebox{\linewidth}{!}{
\begin{tabular}{lllll}
\toprule
 & \textbf{FT} & \textbf{LoRA} & \textbf{ROME} & \textbf{MEMIT}\\
\midrule
\textbf{GPT2-large} & $0.482^*$ & $0.236^*$ & $0.288^*$ & $0.640^*$ \\
\textbf{GPT2-XL} & $0.158^*$ & $0.324^*$ & $0.259^*$ & $0.486^*$ \\
\textbf{GPT-J} & $0.204^*$ & $0.203^*$ & $0.062^*$ & $0.076^*$ \\
\bottomrule
\end{tabular}}
\caption{\label{cf-correlation-table}
\textsc{CounterFact} data Pearson correlation between perplexingness and edit ineffectiveness ($^*$ indicates corresponding entry has $p$-value below 0.05).
}
\end{table}

\paragraph{Correlations are significant} To quantify this relationship, Pearson correlation coefficients are computed and are presented in Table~\ref{cf-correlation-table}. Additionally, to assess the statistical significance of these correlations, $p$-values are calculated. Entries corresponding to $p$-values falling below the significance threshold of 0.05 are marked with $^*$ within the table. 

It is observed that all the coefficients' $p$-values are beneath the 0.05 threshold, thereby indicating a statistically significant correlation between perplexingness and edit ineffectiveness. \textbf{This means that when a model finds new knowledge very perplexing, it is difficult to incorporate this knowledge into the model.} Similarly, a person might be resistant to learning something they find hard to understand.

Furthermore, the analysis reveals that certain scenarios exhibit high Pearson coefficients, such as the application of MEMIT to the GPT-2 large model. This variance could stem from the possibility that different models encode perplexingness in distinct manners and that editing methods may interact with this perplexingness uniquely.

\paragraph{Correlation is in the new knowledge but not the original knowledge} Our analysis specifically focuses on the conditional probabilities of newly introduced knowledge $(s, r, o^*)$, as opposed to the original knowledge $(s, r, o)$ that stored in the language models. Early efforts to evaluate the conditional probabilities of the original knowledge did not show any significant correlation with the editing process, suggesting a mostly arbitrary relationship.

%\subsection{Hierarchical relations}
\subsection{Significantly higher perplexingness of higher hierarchy level knowledge}
Do hierarchical relations affect perplexingness? To investigate the effect of hierarchical relations on ``perplexingness,'' we analyze these two groups in \textsc{HierarchyData}: hypernyms (abstract concepts) and hyponyms (specific concepts). The box plots are included in Appendix D. We conduct $t$-tests for two independent samples to determine if the mean perplexingness of the specific level is statistically lower than that of the abstract level. The results of the $t$-tests are detailed in Table~\ref{hierarchy-preedit-table}, with all values demonstrating statistical significance. Our findings indicate that \textbf{knowledge on a higher hierarchical level (more abstract) is associated with greater perplexingness for the models}. This suggests that hierarchical relations are a factor affecting knowledge perplexingness for language models.

\begin{table}
\centering
\resizebox{.8\linewidth}{!}{
\begin{tabular}{lll}
\toprule 
\textbf{GPT2-Large} & \textbf{GPT2-XL} & \textbf{GPT-J}\\
\midrule 
$0.00728^*$ & $0.00605^*$ & $1.33e-06^*$  \\
\bottomrule
\end{tabular}}
\caption{\label{hierarchy-preedit-table}
Comparative analysis of perplexingness in \textsc{HierarchyData}: $t$-test results for specific vs. abstract level distributions ($*$ indicates corresponding entry has $p$-value below 0.05).
}
\end{table}

\paragraph{Correlations between perplexingness and edit ineffectiveness in \textsc{HierarchyData}}
Next, we aim to determine if the correlation between perplexingness and edit ineffectiveness also holds for the \textsc{HierarchyData} dataset. We employ the same method to analyze \textsc{HierarchyData} as analyzing \textsc{CounterFact}, focusing on the Pearson correlation coefficient between perplexingness and edit ineffectiveness. The right panel of Figure~\ref{gpt2xl_memit_c} provides one of the scatter plots (see Appendix B for other plots), showcasing the application of MEMIT on GPT2-XL. We also calculate the Pearson coefficients, with the results presented in Table~\ref{h-correlation-table}. In this table, $p$-values below 0.05 are marked with $^*$, indicating statistical significance. Our analysis reveals a consistent trend: an increase in perplexingness correlates with poorer efficacy of edits (higher negative log conditional probability). This pattern holds true across all scenarios, except when applying the ROME and MEMIT techniques to the GPT-J model.

\begin{table}
\centering
\resizebox{\linewidth}{!}{
\begin{tabular}{lllll}
\toprule 
 & \textbf{FT} & \textbf{LoRA} & \textbf{ROME} & \textbf{MEMIT}\\
\midrule 
\textbf{GPT2-large} & $0.893^*$ & $0.886^*$ & $0.167^*$ & $0.575^*$ \\
\textbf{GPT2-XL} & $0.860^*$ & $0.856^*$ & $0.148^*$ & $0.381^*$ \\
\textbf{GPT-J} & $0.454^*$ & $0.755^*$ & $0.078$ & $-0.019$ \\
\bottomrule
\end{tabular}}
\caption{\label{h-correlation-table}
\textsc{HierarchyData} Pearson correlation between perplexingness and edit ineffectiveness ($^*$ indicates corresponding entry has $p$-value below 0.05)
}
\end{table}

\subsection{Relationships between hierarchical relations and edit ineffectiveness}
\begin{figure}[t]
\vskip 0.2in
\begin{center}
\centerline{\includegraphics[width=\columnwidth]{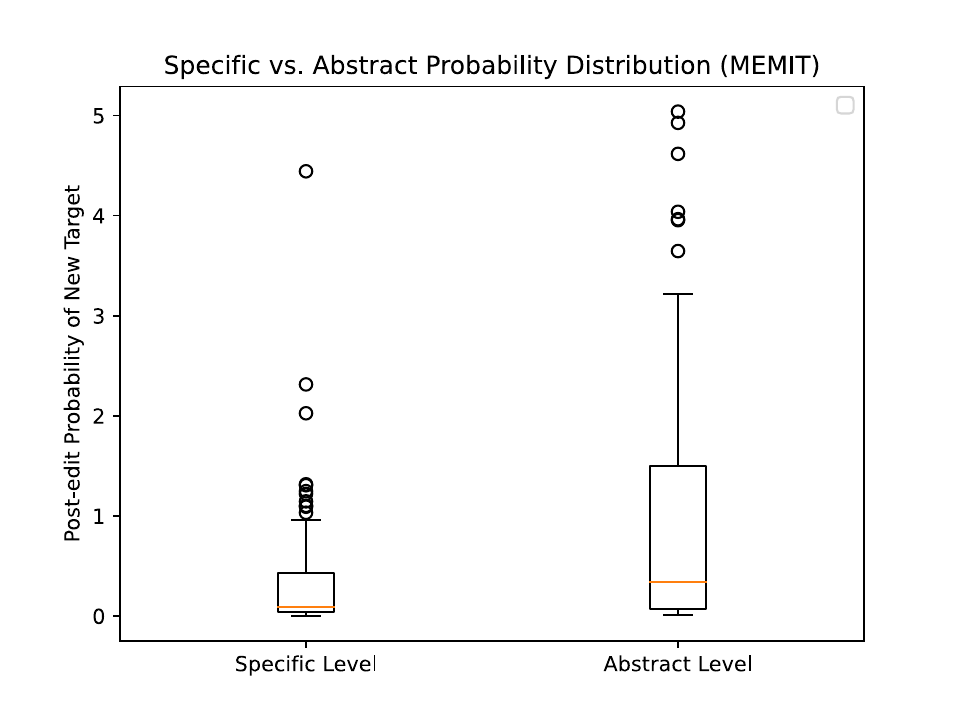}}
\caption{The post-edit probability (lower probability means higher edit efficacy) of editing GPT2-XL with MEMIT on specific vs. abstract knowledge in the \textsc{HierarchyData}.}
\label{gpt2xl_memit_h}
\end{center}
\vskip -0.2in
\end{figure}

% \TODO{not significant}
Additionally, we want to determine if hierarchical relations within the knowledge ultimately affect the edit ineffectiveness. Box plots (see Appendix C) are constructed to visually compare the ineffectiveness across the two hierarchical levels. Figure~\ref{gpt2xl_memit_h} shows one of the examples. Furthermore, we conduct $t$-tests on two independent samples to determine whether the mean of the specific level distribution is significantly lower than that of the abstract level distribution. The $p$-values obtained are documented in Table~\ref{hierarchy-table}. This finding underscores a markedly lower efficacy in editing knowledge at higher hierarchical levels (more abstract knowledge). Significantly, this discrepancy indicates that hierarchical relationships profoundly affect the efficacy of specific editing techniques, like ROME and MEMIT, when applied to particular models, such as GPT2-Large and GPT2-XL. For fine-tuning and LoRA, the results do not appear to be significant, possibly because these methods can address knowledge at different hierarchical levels similarly. But, how about GPT-J?

\begin{figure}[t]
\vskip 0.2in
\begin{center}
\centerline{\includegraphics[width=\columnwidth]{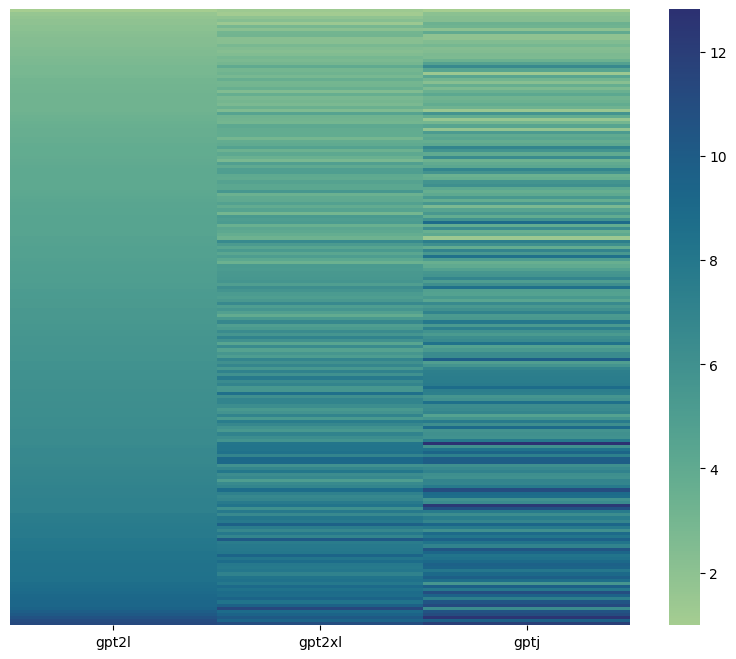}}
\caption{\textbf{Same knowledge perplexingness in different models (\textsc{HierarchyData}).} Each line represents a piece of knowledge from \textsc{HierarchyData}, sorted by perplexingness in the GPT-2L model. We observe that GPT-J appears darker in the heatmap, suggesting it finds the same knowledge less perplexing.}
\label{sameheatmap}
\end{center}
\vskip -0.2in
\end{figure}

\paragraph{GPT-J can understand perplexing knowledge better}

From the previous experiment, we observe that GPT-J did not show any difference in edit efficacy when editing higher hierarchy and lower hierarchy knowledge. To determine if GPT-J finds the same knowledge less perplexing compared to GPT-2L and GPT-2XL, we generated a heatmap of each knowledge's perplexingness in the \textsc{HierarchyData} for each model, as shown in Figure~\ref{sameheatmap}. Each line represents a piece of knowledge in the \textsc{HierarchyData}, sorted by perplexingness in the GPT-2L model. We observed that GPT-J appears darker in the heatmap, indicating it finds the same knowledge less perplexing.

To assess the statistical significance of this observation, we conduct paired $t$-tests comparing the perplexingness values of GPT-J to those of GPT-2L and GPT-2XL. The resulting $p$-values were $5.71e-9$ and $6.84e-7$, respectively, indicating a very significant difference. This suggests that GPT-J indeed finds the same knowledge less perplexing than GPT-2L and GPT-2XL, implying that GPT-J is more receptive to learning new things. Additionally, this means GPT-J can learn more beyond hierarchical relationships, and various factors will influence its edit efficacy.

\begin{table}
\centering
\resizebox{\linewidth}{!}{
\begin{tabular}{lllll}
\toprule 
 & \textbf{FT} & \textbf{LoRA} & \textbf{ROME} & \textbf{MEMIT}\\
\midrule
\textbf{GPT2-large} & $0.970$ & $0.989$ & $0.113$ & $3.41e-8^*$ \\
\textbf{GPT2-XL} & $0.972$ & $0.958$ & $0.0286^*$ & $8.14e-6^*$ \\
\textbf{GPT-J} & $0.865$ & $0.770$ & $0.317$ & $0.976$ \\
\bottomrule
\end{tabular}}
\caption{\label{hierarchy-table}
Comparative analysis of ineffectiveness in \textsc{HierarchyData}: $t$-test results for specific vs. abstract level distributions ($*$ indicates corresponding entry has $p$-value below 0.05).
}
\end{table}

% \paragraph{Lack of Significant Findings Across Knowledge Categories} Besides hierarchical relations, we also try to find if categories of knowledge would affect perplexingness. We attempt to categorize the data based on the types of knowledge; however, this method does not yield any significant insights related to perplexingness.

\section{Discussion}

% different topics

\paragraph{Do different models have different mechanisms of saving perplexing knowledge?}
Our experimental results reveal intriguing variations in how different models handle perplexing knowledge, particularly in the context of editing. Specifically, the application of ROME and MEMIT to GPT-J exhibits a notably low Pearson correlation between perplexingness and editing ineffectiveness. Moreover, within the \textsc{HierarchyData} context, these correlations appear insignificant. Additionally, the influence of hierarchical relations on the editing ineffectiveness of ROME and MEMIT when applied to GPT-J seems negligible. This suggests that GPT-J may employ a unique mechanism for storing and processing different hierarchy-level knowledge compared to other models. These differences highlight the need to comprehend each model's unique architecture and methods for handling perplexing concepts, suggesting a move towards tailored editing strategies.

\paragraph{Why should more abstract knowledge be harder to edit?} An intuition is that when editing towards a hypernym (``animal'' $\rightarrow$ ``plant''), it is assumed that the hyponym (``cat'' $\rightarrow$ ``plant'') is edited as well, making the edit of hypernym inherently harder. Yet, the dependent knowledge is usually not edited, for popular editing methods \citep{li2023evaluating}.

\paragraph{Are there other factors that may influence the perplexingness?}
The investigation into the responsiveness of different editing techniques to perplexing knowledge reveals that FT and LoRA are seemingly unaffected by the hierarchical structure of knowledge. Notably, there exists a pronounced correlation between perplexingness and the ineffectiveness of edits. This suggests that while FT and LoRA are adept at navigating the hierarchical relationships among words, they falter when addressing the inherent perplexingness present within the knowledge. This observation leads to the hypothesis that additional factors, beyond hierarchical complexity, play a pivotal role in influencing perplexingness when employing FT and LoRA for knowledge editing.

\begin{figure}[t]
    \centering 
    \includegraphics[width=\columnwidth]{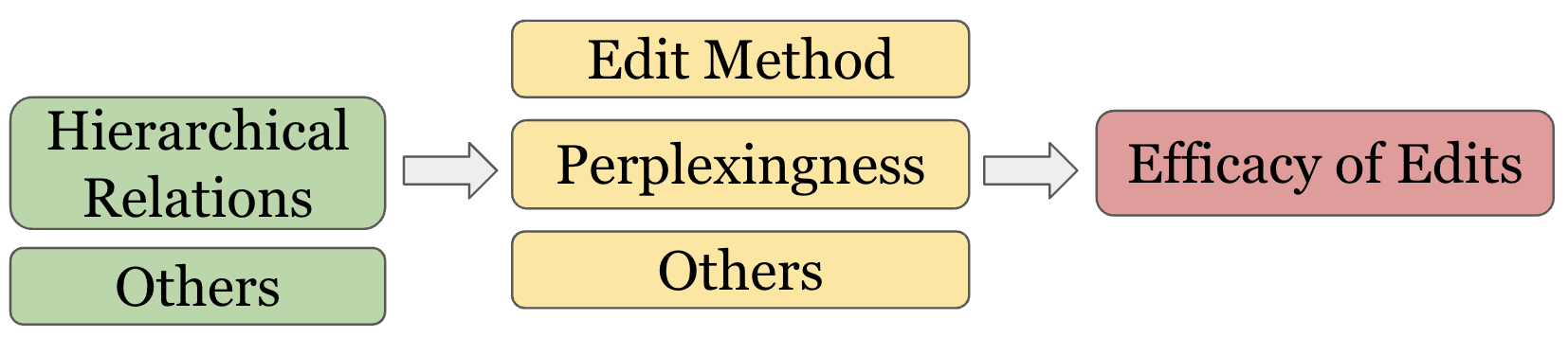}
    \caption{\textbf{Factors that influence the edit efficacy.} In this paper, we explore how hierarchical relations influence perplexingness and how perplexingness, in turn, affects the efficacy of edits. However, other factors may also impact the efficacy of edits, such as the choice of edit methods or models. Additionally, there may be other factors contributing to perplexingness, such as fine-grained or ambiguous knowledge. Investigating these aspects is beyond the scope of this work and will be addressed in future research.}
    \label{fig:edit-efficacy-reasons}
\end{figure}

\paragraph{More understanding of model editing}
The impact of perplexingness on the ineffectiveness of various editing methodologies can vary significantly. Moreover, the manner in which different models interpret, process, and encode the perplexingness of knowledge also differs. This suggests a complex interplay between the editing methods used and the intrinsic mechanisms of the models, as illustrated by Figure \ref{fig:edit-efficacy-reasons}, underscoring the need for a nuanced understanding of both to optimize knowledge editing strategies. Other factors beyond hierarchical complexity may also contribute to the perplexingness of knowledge. For instance, knowledge with significant semantic overlap with other concepts can introduce perplexingness by creating competing or conflicting representations. Similarly, fine-grained or highly specific knowledge, as well as ambiguous knowledge with multiple possible interpretations, may further increase perplexingness.

\paragraph{Recommendations to future model editors}
Future model editing efforts should pay attention to understanding the nature of the knowledge being edited, particularly its level of perplexingness. To aid in this endeavor, we have introduced a hierarchy dataset designed to facilitate it. It is crucial to ensure that editing methods are versatile and effective across a diverse range of data types. Moreover, adopting different editing approaches tailored to the specificities of each model can significantly enhance the success of edits. When editing hierarchy knowledge, we can try to use edit methods like fine-tuning or LoRA. It may dismiss the influence of hierarchy data. Also, we should pay attention to the side effects of knowledge edit. 

% \TODO{future work and possible solution}

\section{Conclusion}
Our investigation into knowledge editing in LLMs reveals a fundamental challenge: the more perplexing a piece of knowledge is to an LLM, the more resistant it becomes to modification through existing editing methods. Through comprehensive analysis using both the \textsc{CounterFact} dataset and our newly developed \textsc{HierarchyData}, we demonstrate that abstract concepts (hypernyms) are inherently more perplexing to LLMs than their specific counterparts (hyponyms), leading to lower editing efficacy. These findings not only highlight a previously unexplored aspect of model editing technology but also provide crucial insights for developing more sophisticated editing methodologies that can effectively handle knowledge across different levels of conceptual abstraction.

\section{Limitation}
In this paper, we focus on a short hierarchy chain to facilitate the comparison between higher and lower hierarchy levels. Future works can explore longer hierarchy chains. 
The experiment can be scaled up, including the use of larger models and larger datasets. Additional types of evaluation can be applied. For instance, we could ask language model-specific questions to determine if the knowledge has actually been edited. However, this approach is very labor-intensive and was not implemented in this study.

\section{Acknowledgements}
A previous version of this document contained a hidden prompt entered by Z Zhu without knowledge of -- or consent by -- his co-authors. This version does not contain the prompt.

\bibliography{custom}

\appendix

% \section{Example Appendix}
% \label{sec:appendix}

\section{Correlation of perplexingness and efficacy in \textsc{CounterFact}}
\label{sec:appendix-correlation-counterfact}

\begin{figure}[h]
\begin{center}
% \framebox[4.0in]{$\;$}
\includegraphics[width=6cm]{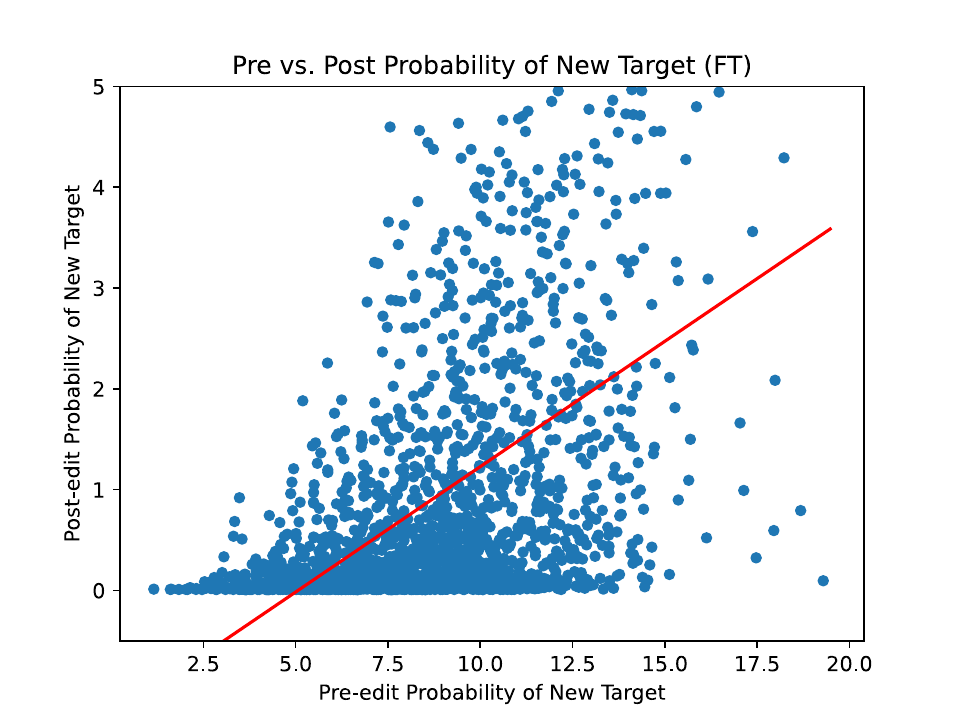}
\includegraphics[width=6cm]{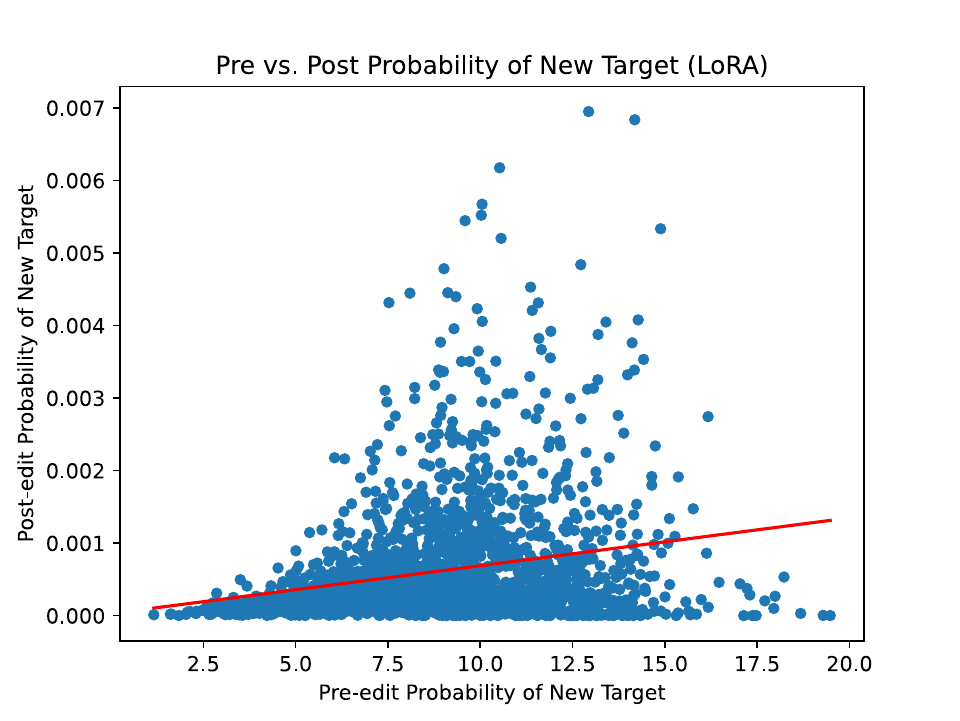}
\includegraphics[width=6cm]{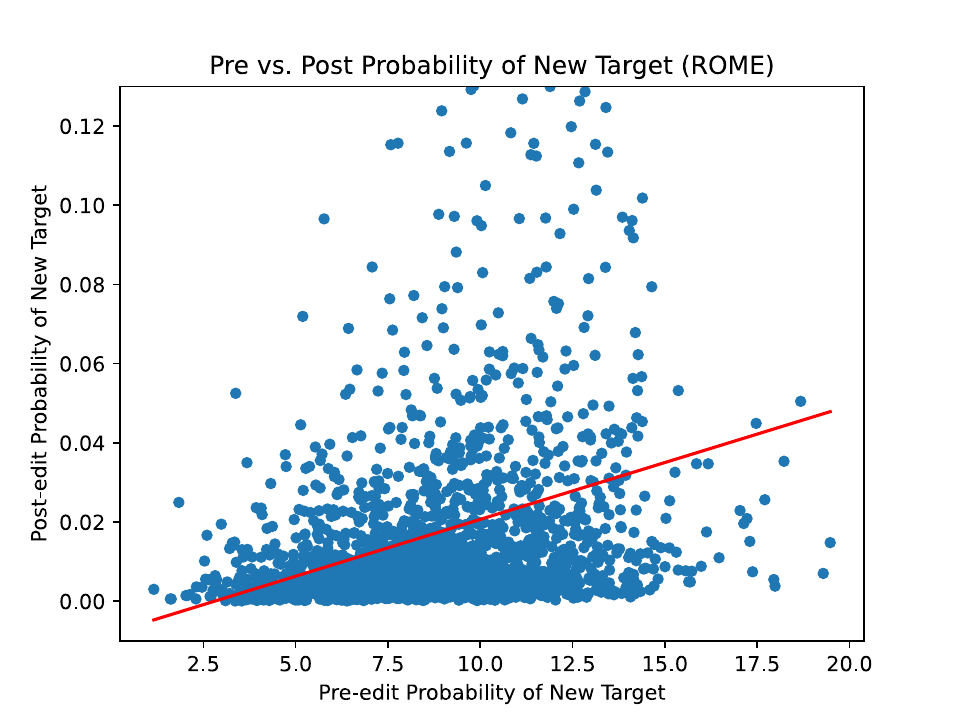}
\includegraphics[width=6cm]{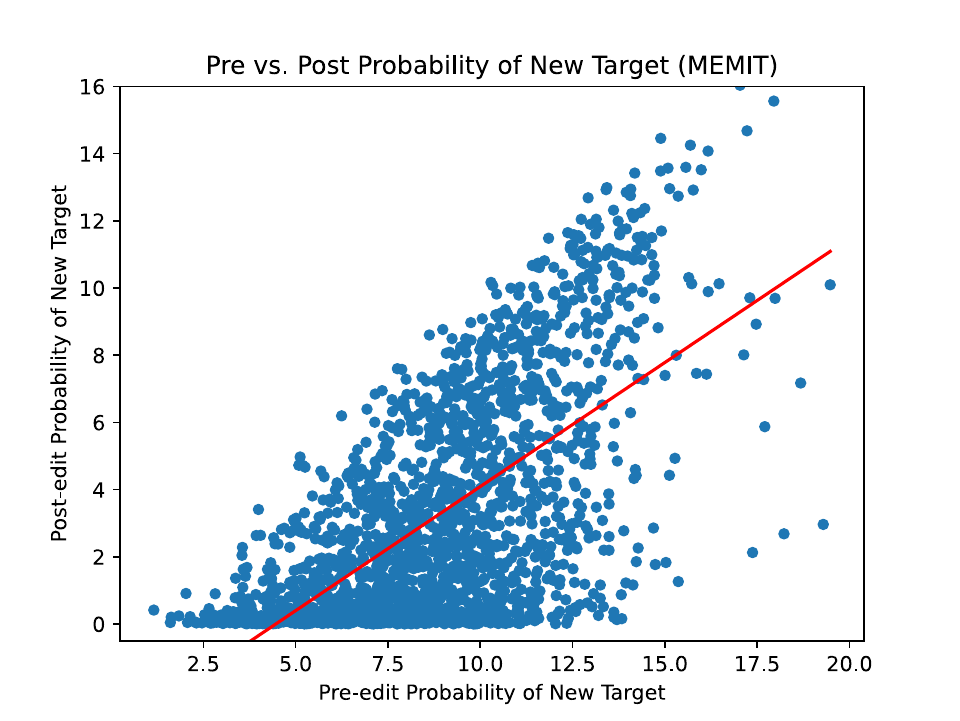}
% \fbox{\rule[-.5cm]{0cm}{4cm} \rule[-.5cm]{4cm}{0cm}}
\end{center}
\caption{Pre vs. post probability of new knowledge (\textsc{CounterFact}) on GPT2-Large using a. FT (first) b. LoRA (second) c. ROME (third) d. MEMIT (fourth). }
\label{all_gpt2l_cf}
\end{figure}

\begin{figure}[h]
\begin{center}
% \framebox[4.0in]{$\;$}
\includegraphics[width=6cm]{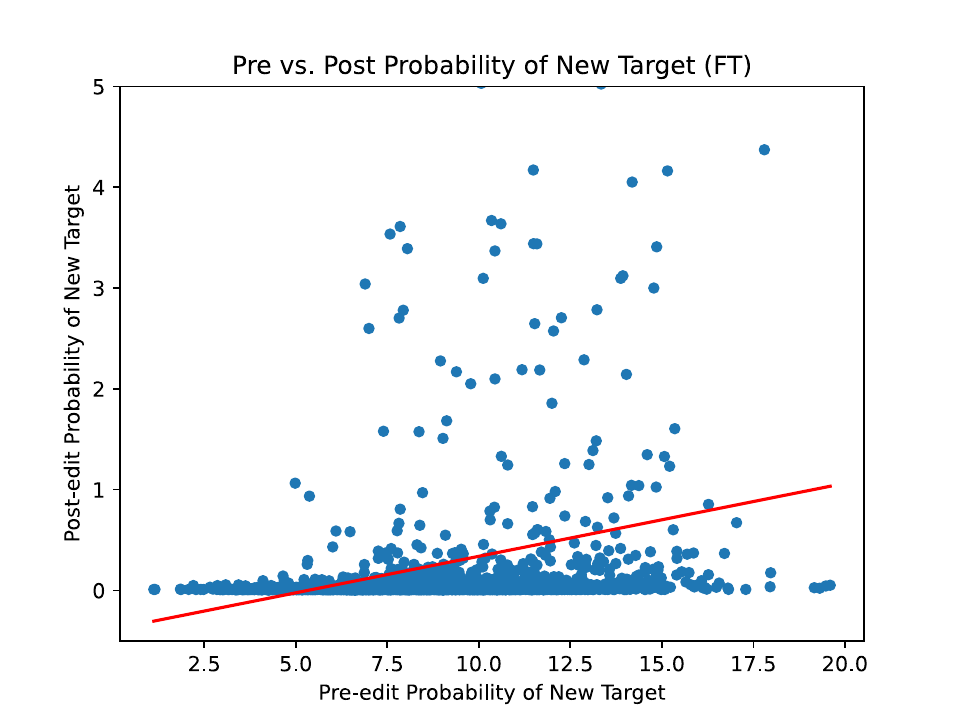}
\includegraphics[width=6cm]{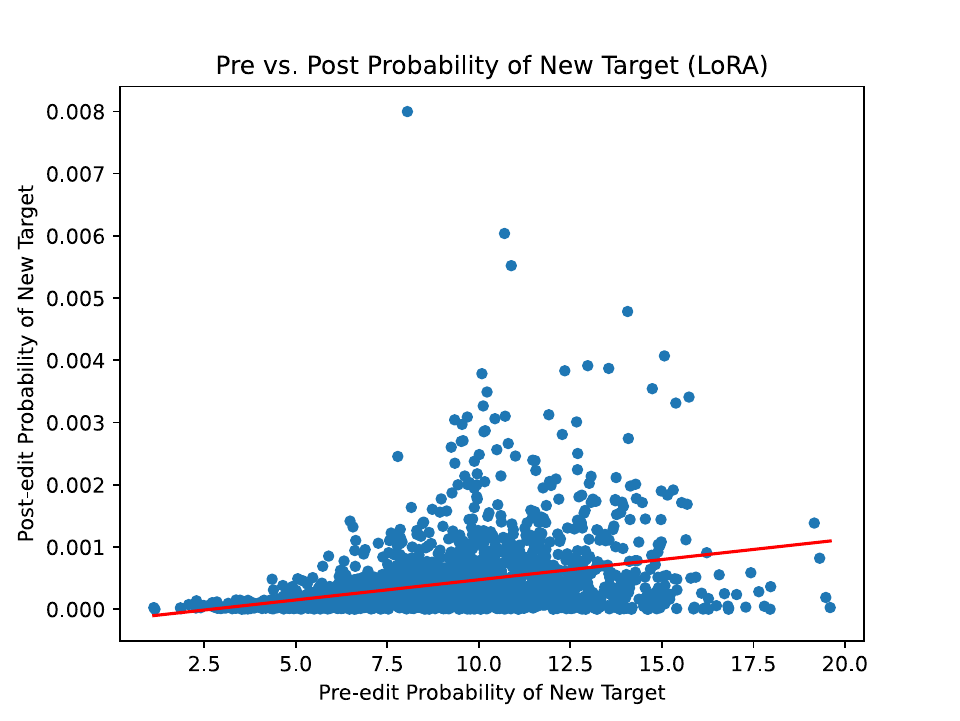}
\includegraphics[width=6cm]{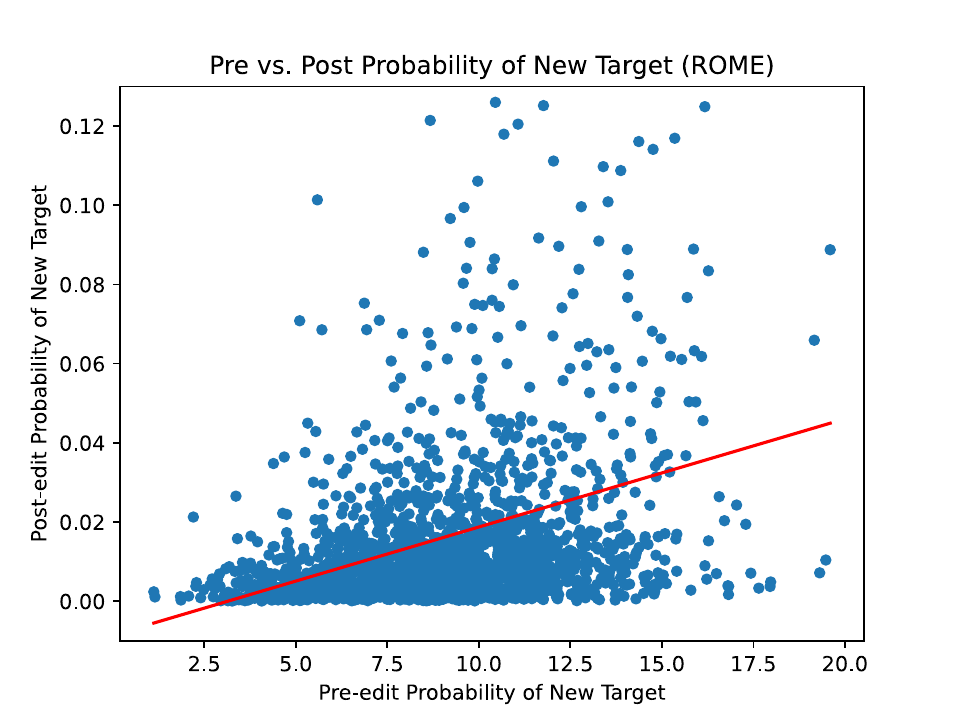}
\includegraphics[width=6cm]{gpt2xl_memit_cf.pdf}
% \fbox{\rule[-.5cm]{0cm}{4cm} \rule[-.5cm]{4cm}{0cm}}
\end{center}
\caption{Pre vs. post probability of new knowledge (\textsc{CounterFact}) on GPT2-XL using a. FT (first) b. LoRA (second) c. ROME (third) d. MEMIT (fourth). }
\label{all_gpt2xl_cf}
\end{figure}

\begin{figure}[h]
\begin{center}
% \framebox[4.0in]{$\;$}
\includegraphics[width=6cm]{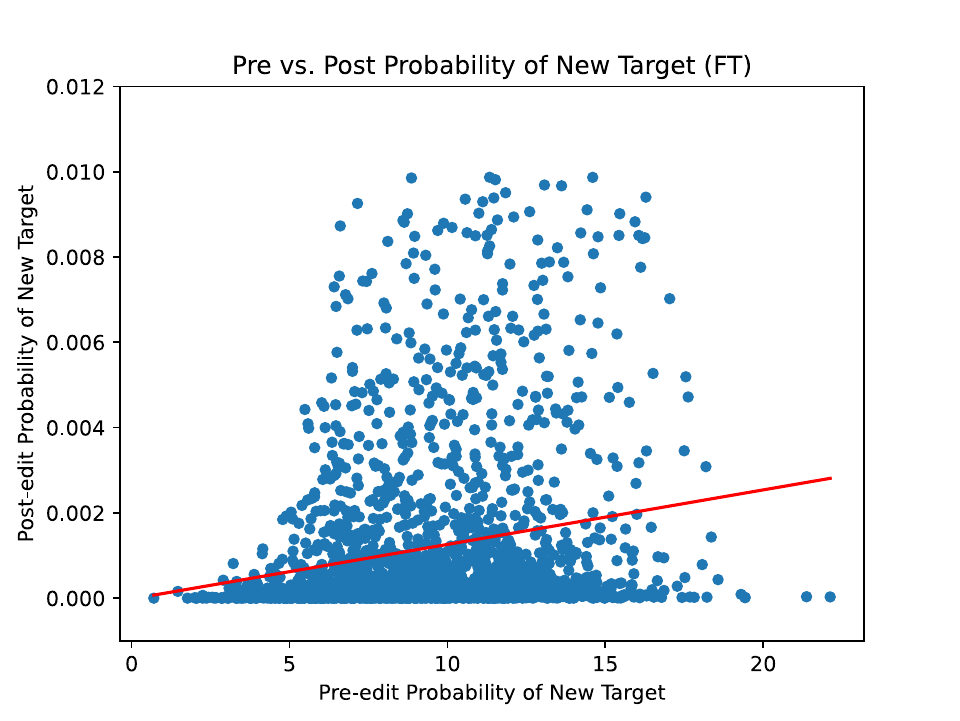}
\includegraphics[width=6cm]{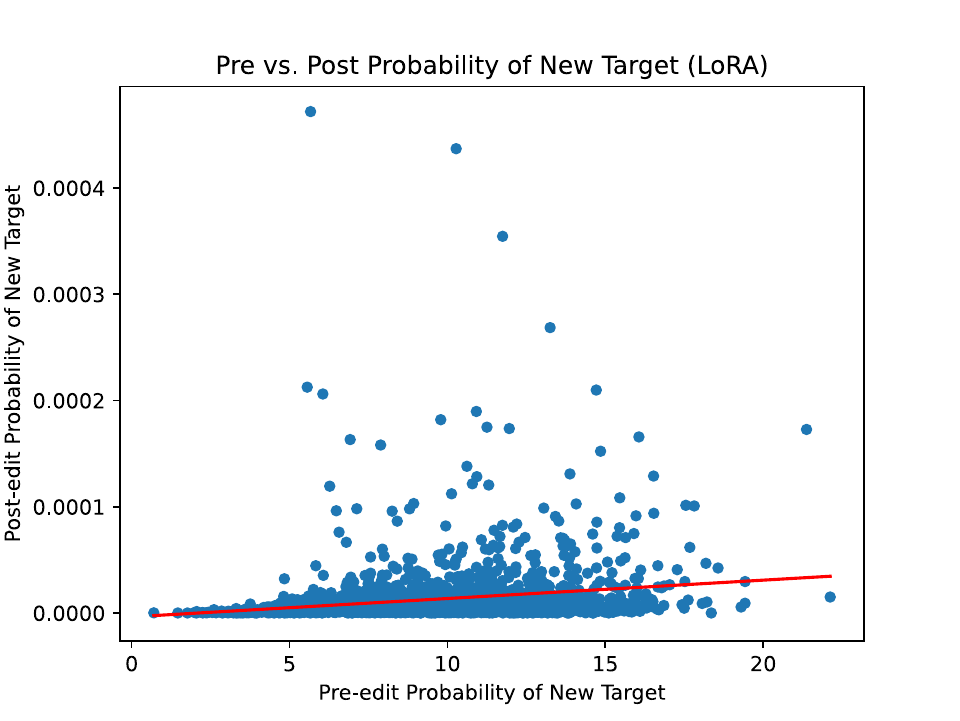}
\includegraphics[width=6cm]{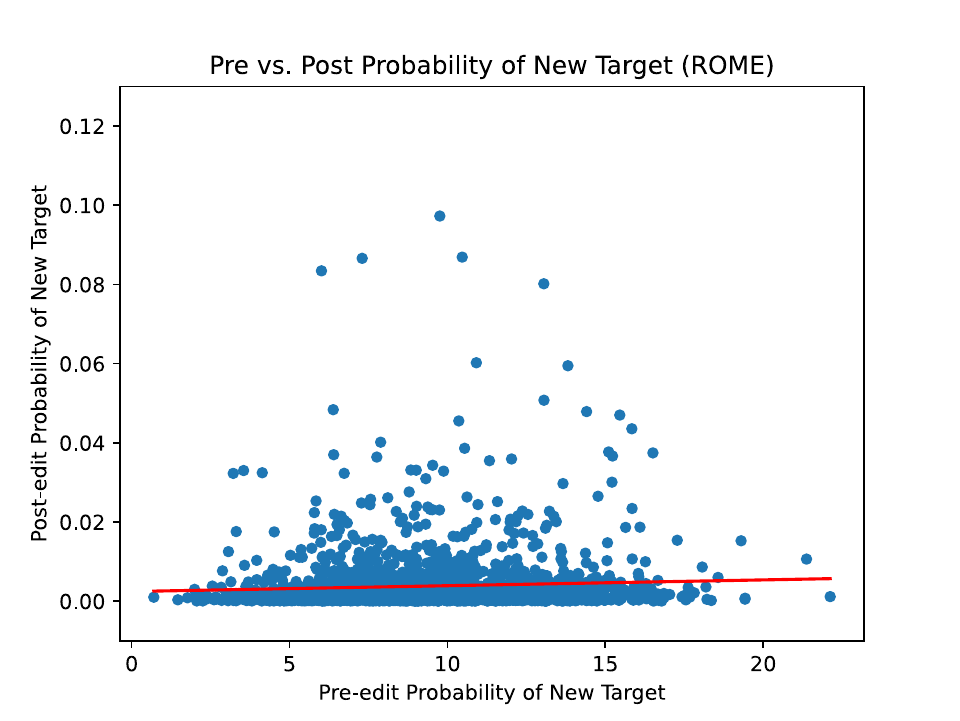}
\includegraphics[width=6cm]{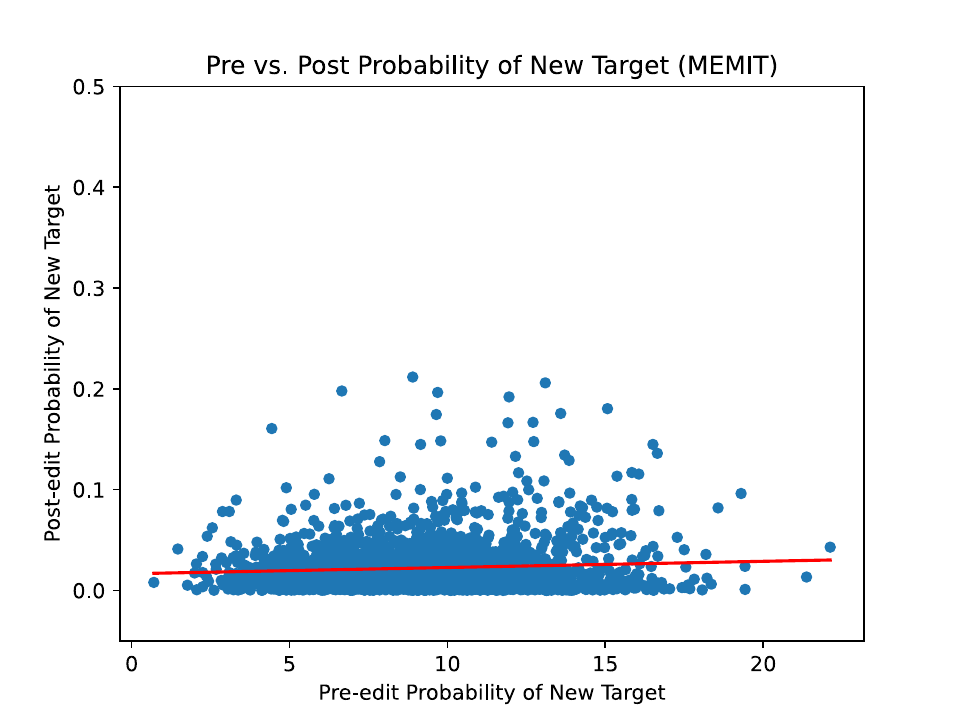}
% \fbox{\rule[-.5cm]{0cm}{4cm} \rule[-.5cm]{4cm}{0cm}}
\end{center}
\caption{Pre vs. post probability of new knowledge (\textsc{CounterFact}) on GPT-J(6B) using a. FT (first) b. LoRA (second) c. ROME (third) d. MEMIT (fourth). }
\label{all_gptj_cf}
\end{figure}

We plot the perplexingness (pre-edit probabilities of the new target) against the efficacy (post-edit probabilities of the new target) to visually analyze their relationship. This analysis is conducted using the first 2000 groupings from the \textsc{CounterFact} dataset. Figure~\ref{all_gpt2l_cf} displays the scatter plot for editing methods applied to GPT2-Large. Similarly, Figure~\ref{all_gpt2xl_cf} presents the scatter plot for methods used on GPT2-XL, and Figure~\ref{all_gptj_cf} illustrates the scatter plot for edits performed on GPT-J(6B).

\section{Correlation of perplexingness and efficacy in \textsc{HierarchyData}}
\label{sec:appendix-correlation-hierarchydata}

\begin{figure}[h]
\begin{center}
% \framebox[4.0in]{$\;$}
\includegraphics[width=6cm]{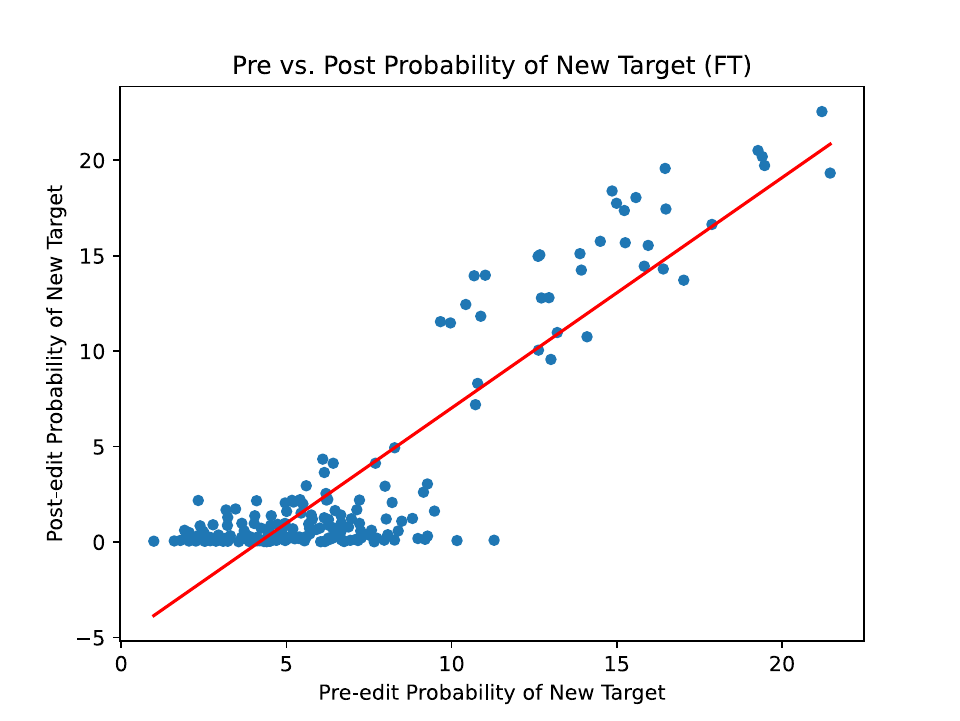}
\includegraphics[width=6cm]{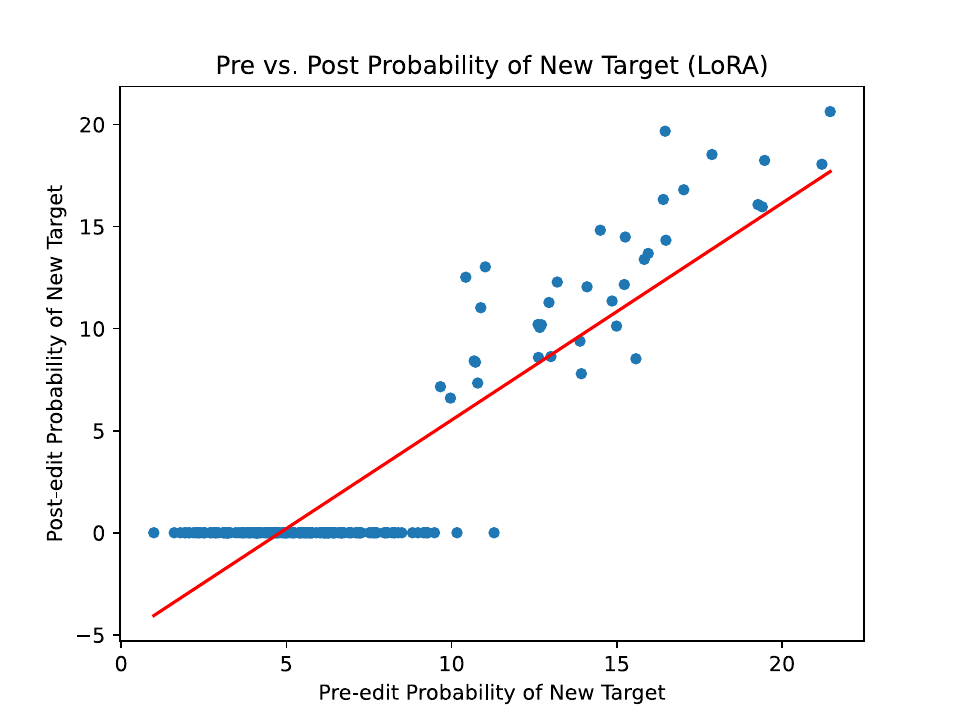}
\includegraphics[width=6cm]{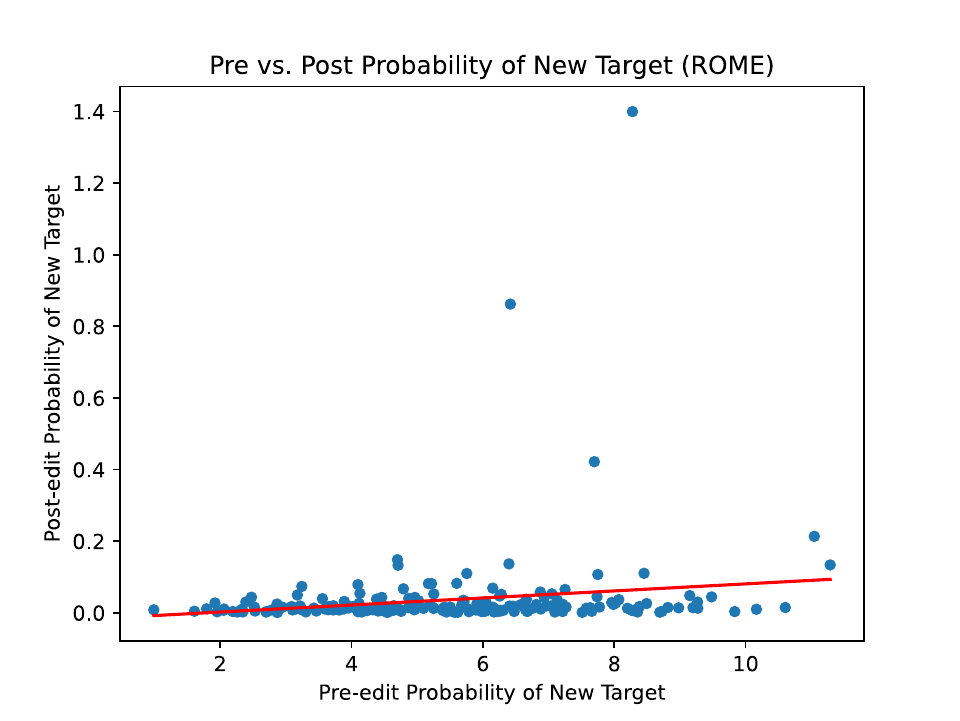}
\includegraphics[width=6cm]{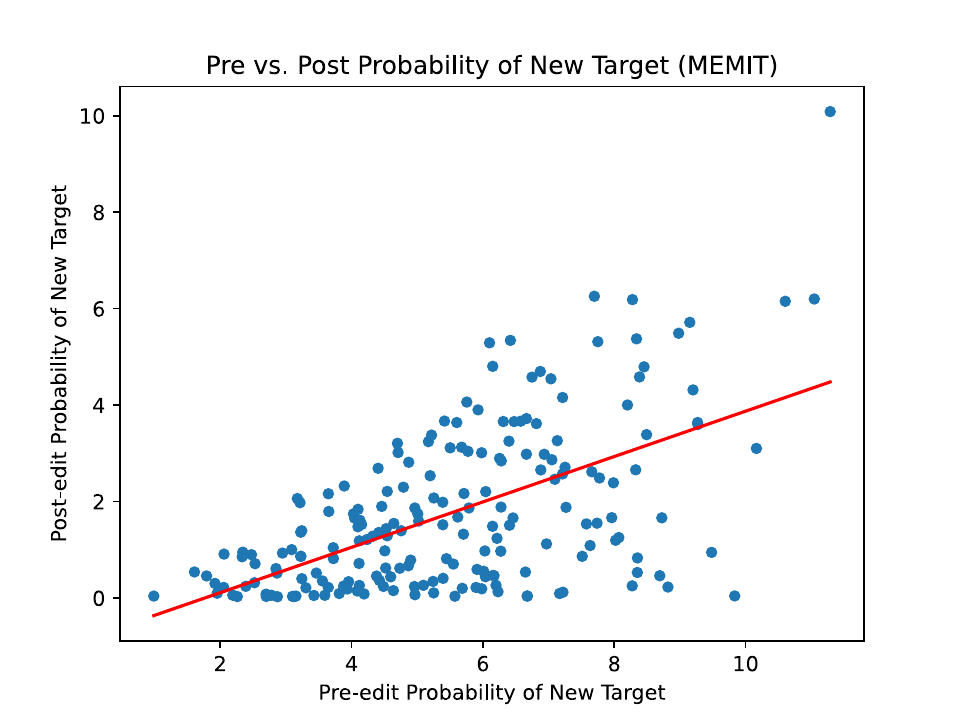}
% \fbox{\rule[-.5cm]{0cm}{4cm} \rule[-.5cm]{4cm}{0cm}}
\end{center}
\caption{Pre vs. post probability of new knowledge (\textsc{HierarchyData}) on GPT2-Large using a. FT (first) b. LoRA (second) c. ROME (third) d. MEMIT (fourth). }
\label{all_gpt2l_hc}
\end{figure}

\begin{figure}[h]
\begin{center}
% \framebox[4.0in]{$\;$}
\includegraphics[width=6cm]{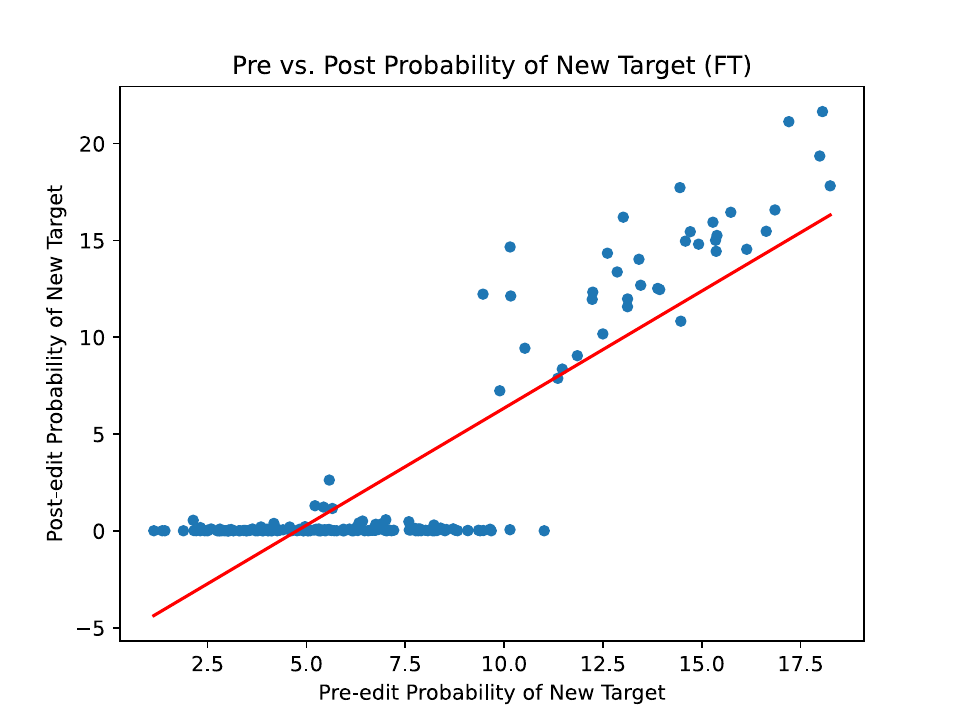}
\includegraphics[width=6cm]{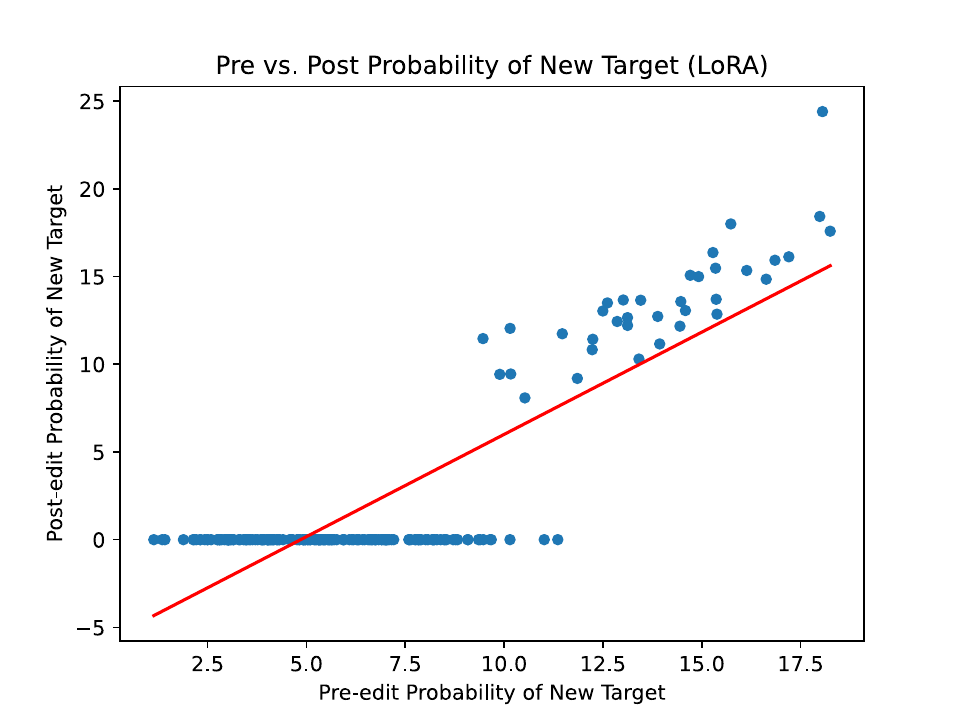}
\includegraphics[width=6cm]{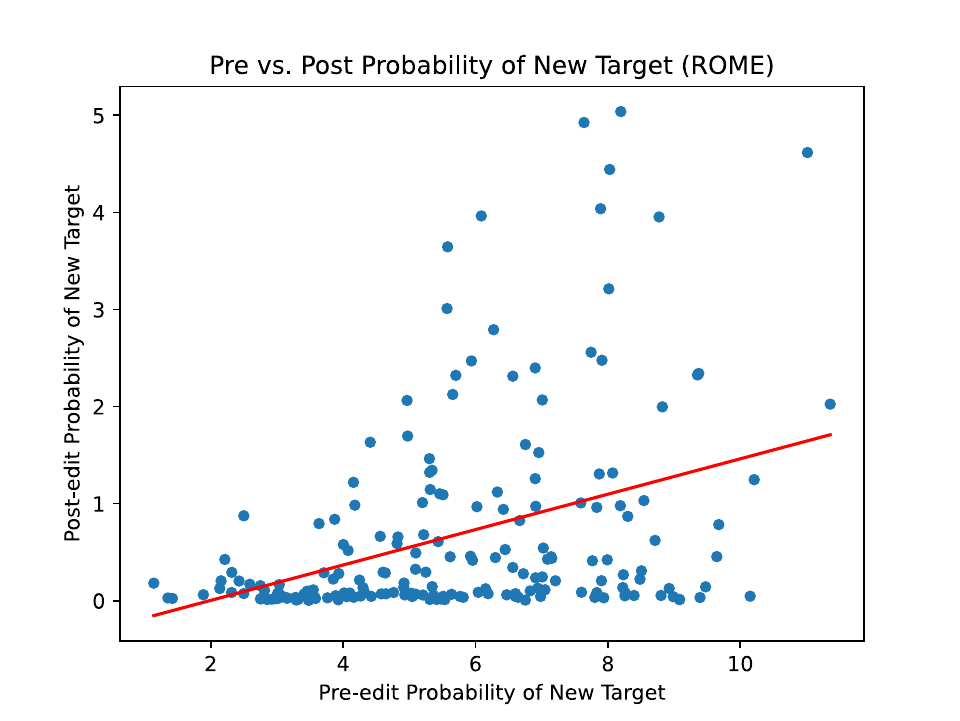}
\includegraphics[width=6cm]{gpt2xl_memit_hc.pdf}
% \fbox{\rule[-.5cm]{0cm}{4cm} \rule[-.5cm]{4cm}{0cm}}
\end{center}
\caption{Pre vs. post probability of new knowledge (\textsc{HierarchyData}) on GPT2-XL using a. FT (first) b. LoRA (second) c. ROME (third) d. MEMIT (fourth). }
\label{all_gpt2xl_hc}
\end{figure}

\begin{figure}[h]
\begin{center}
% \framebox[4.0in]{$\;$}
\includegraphics[width=6cm]{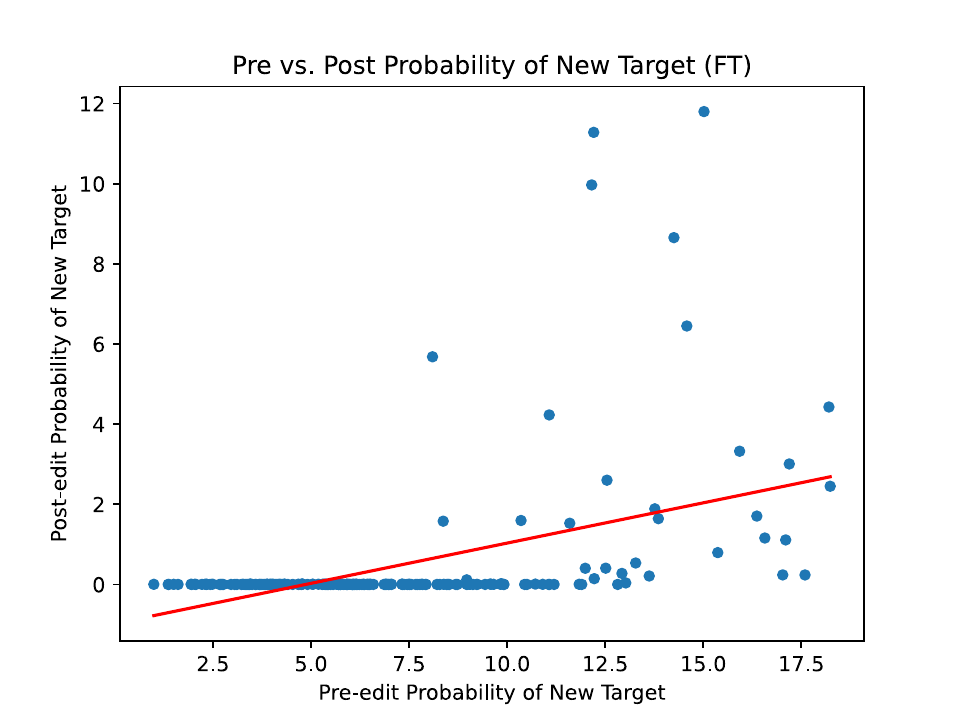}
\includegraphics[width=6cm]{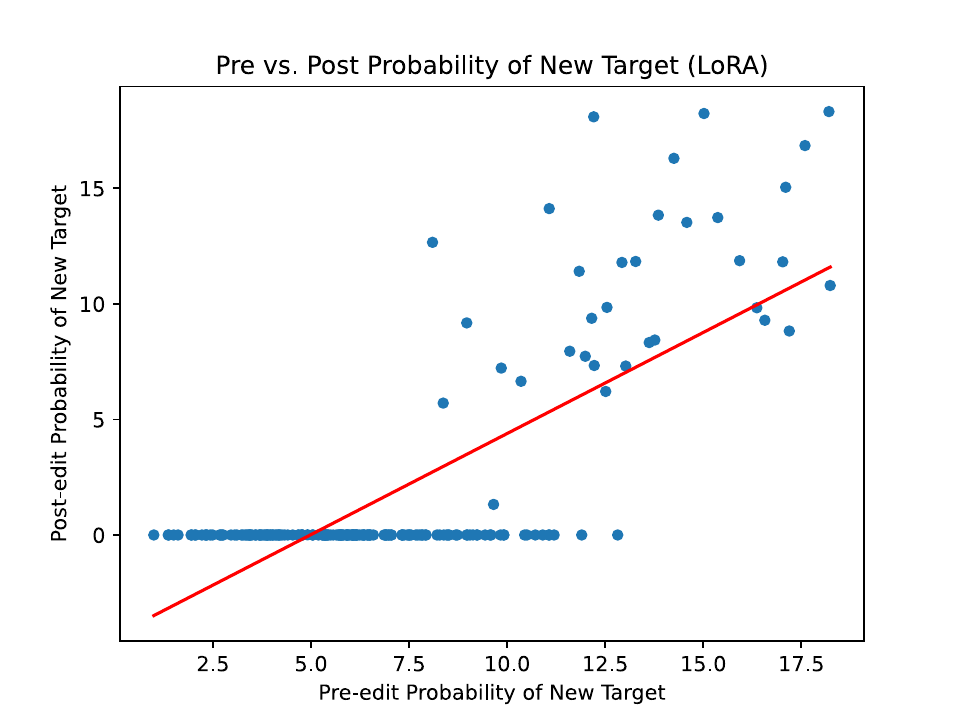}
\includegraphics[width=6cm]{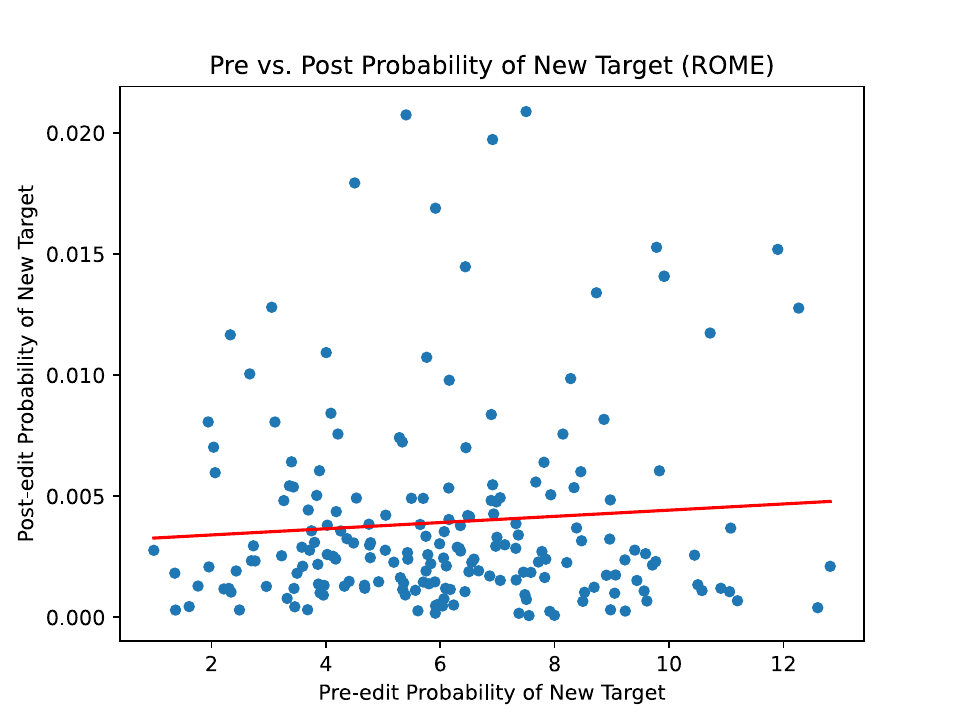}
\includegraphics[width=6cm]{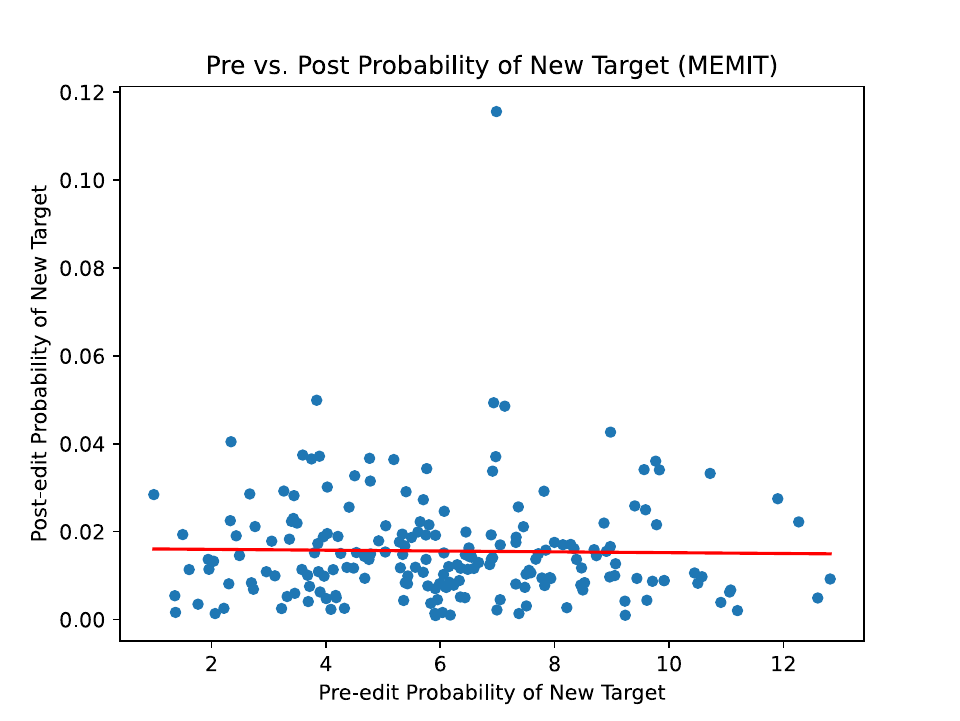}
% \fbox{\rule[-.5cm]{0cm}{4cm} \rule[-.5cm]{4cm}{0cm}}
\end{center}
\caption{Pre vs. post probability of new knowledge (\textsc{HierarchyData}) on GPT-J(6B) using a. FT (first) b. LoRA (second) c. ROME (third) d. MEMIT (fourth). }
\label{all_gptj_hc}
\end{figure}

To visually explore the relationship between perplexingness and editing efficacy, we plot these dimensions against each other using 198 groupings from the \textsc{HierarchyData} dataset. Figure~\ref{all_gpt2l_hc} shows the scatter plot highlighting the effects of editing methods on the GPT2-Large model. Likewise, Figure~\ref{all_gpt2xl_hc} demonstrates the scatter plot for the GPT2-XL model, and Figure~\ref{all_gptj_hc} displays the scatter plot for edits on the GPT-J(6B) model, providing a clear visual representation of how perplexingness correlates with the efficacy of knowledge edits across different models.

\section{Specific vs. Abstract Probability Distribution in \textsc{HierarchyData}}
\label{sec:appendix-probability-hierarchidata}

\begin{figure}[h]
\begin{center}
% \framebox[4.0in]{$\;$}
\includegraphics[width=6cm]{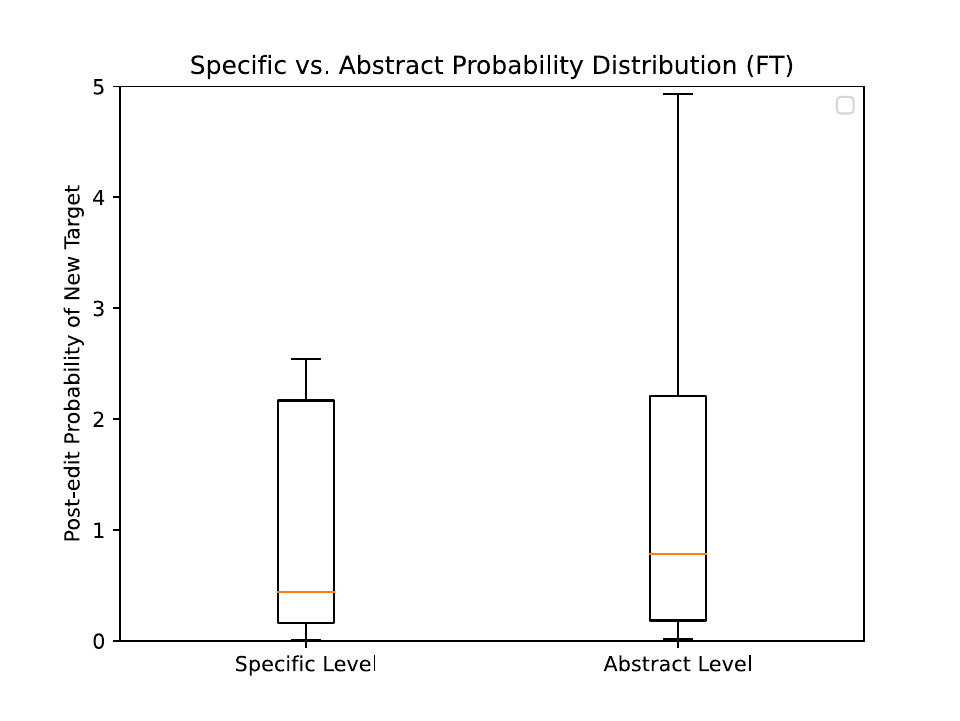}
\includegraphics[width=6cm]{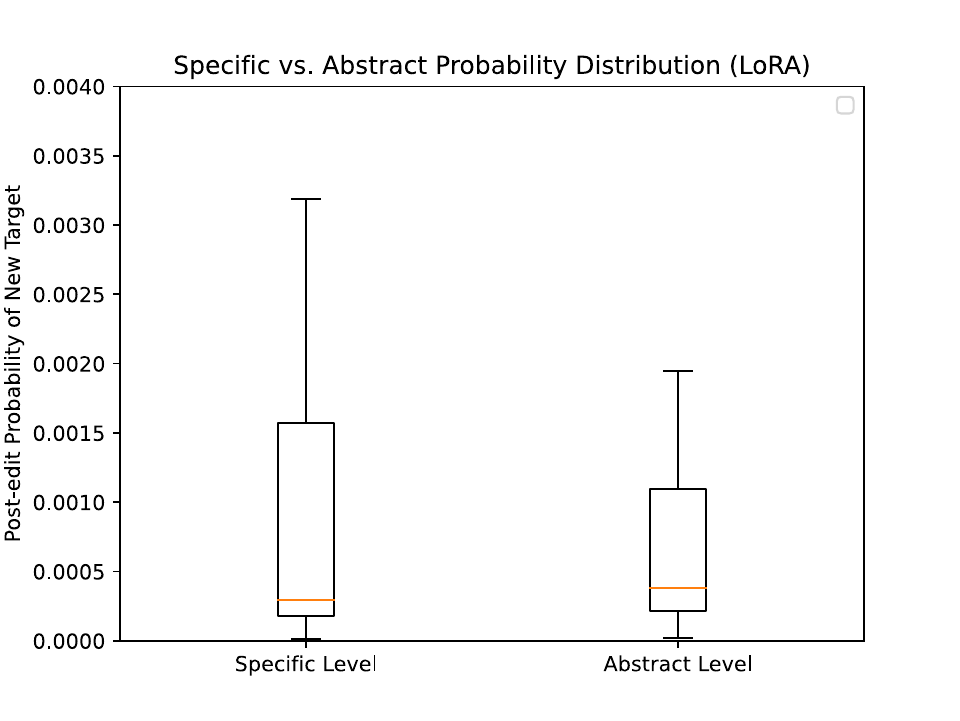}
\includegraphics[width=6cm]{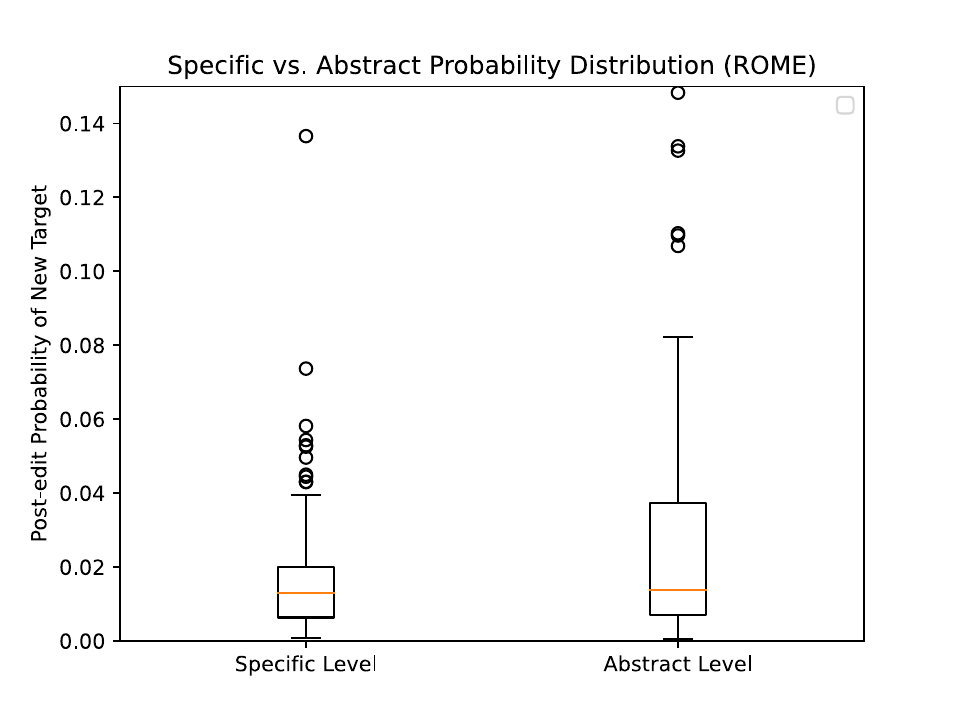}
\includegraphics[width=6cm]{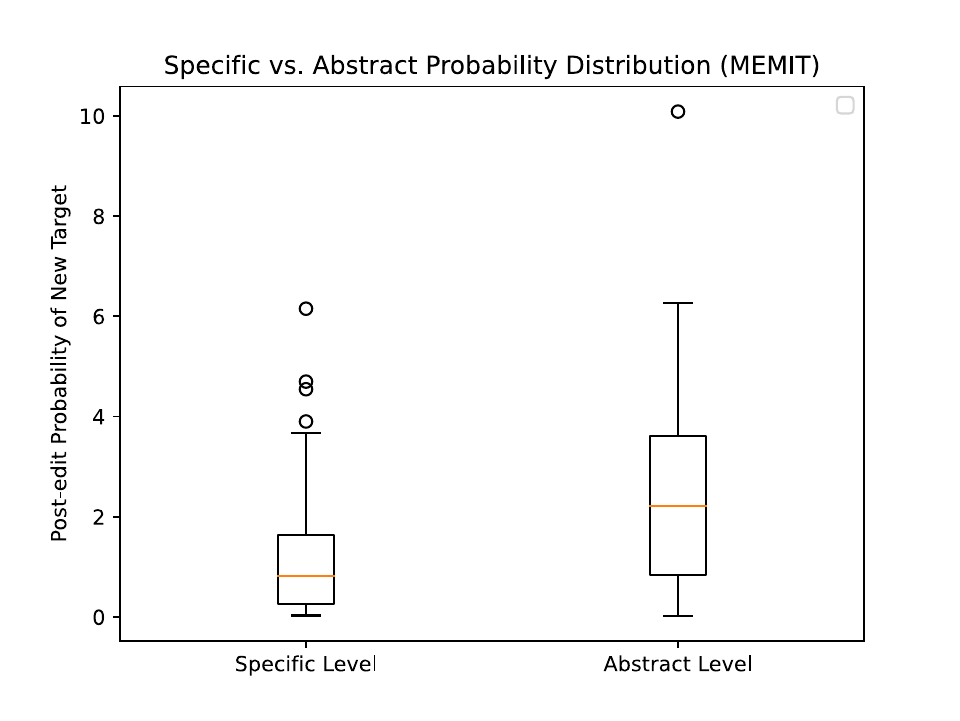}
% \fbox{\rule[-.5cm]{0cm}{4cm} \rule[-.5cm]{4cm}{0cm}}
\end{center}
\caption{Specific vs. abstract probability distribution (\textsc{HierarchyData}) on GPT2-Large using a. FT (first) b. LoRA (second) c. ROME (third) d. MEMIT (fourth). }
\label{all_gpt2l_h}
\end{figure}

\begin{figure}[h]
\begin{center}
% \framebox[4.0in]{$\;$}
\includegraphics[width=6cm]{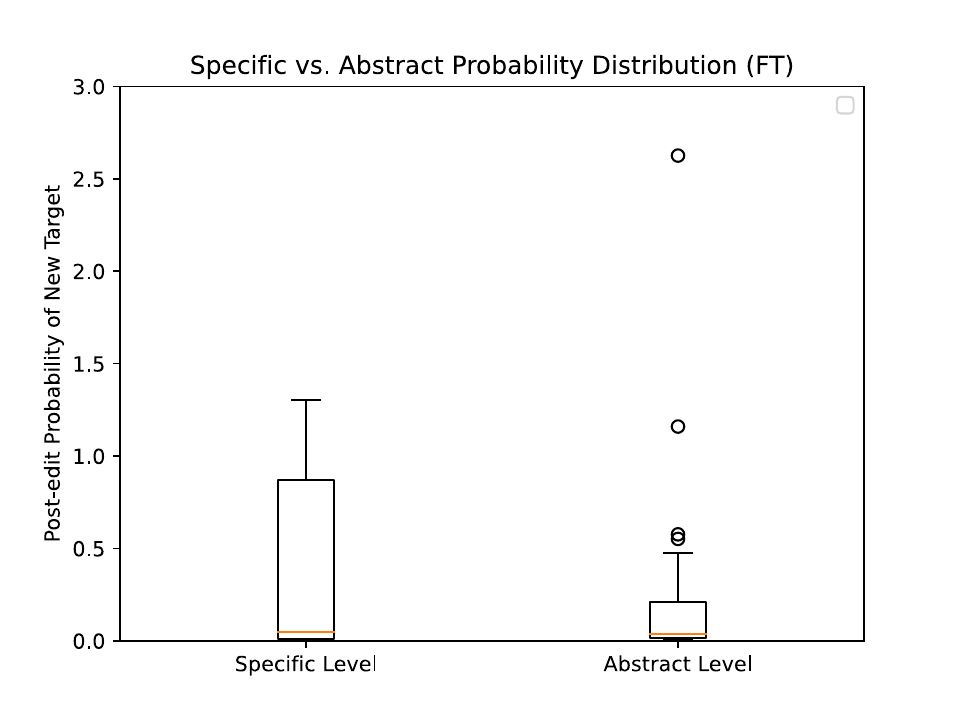}
\includegraphics[width=6cm]{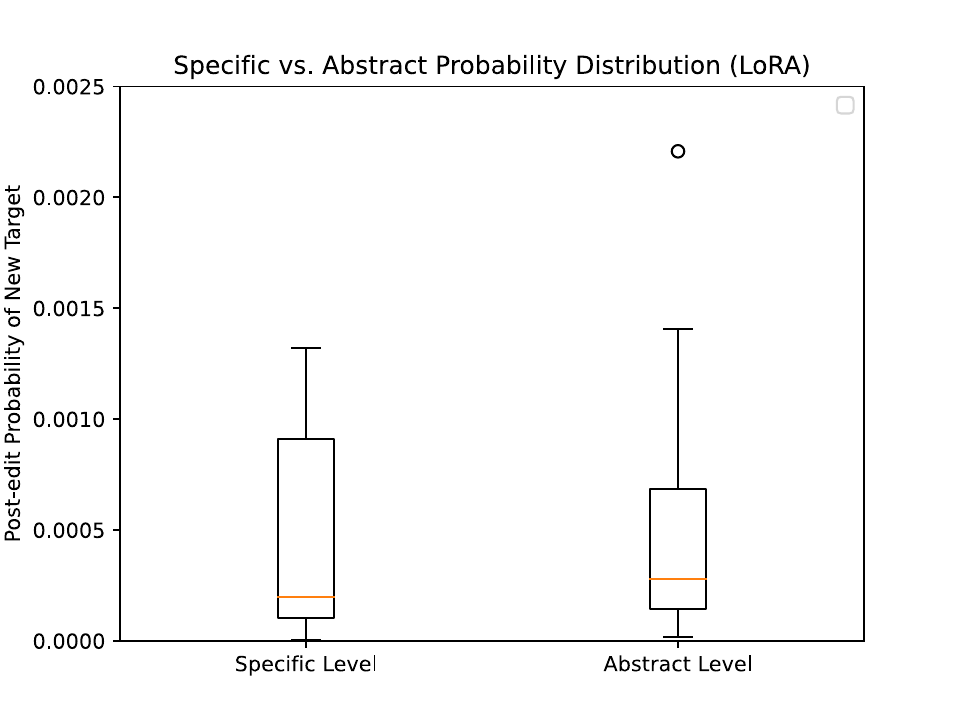}
\includegraphics[width=6cm]{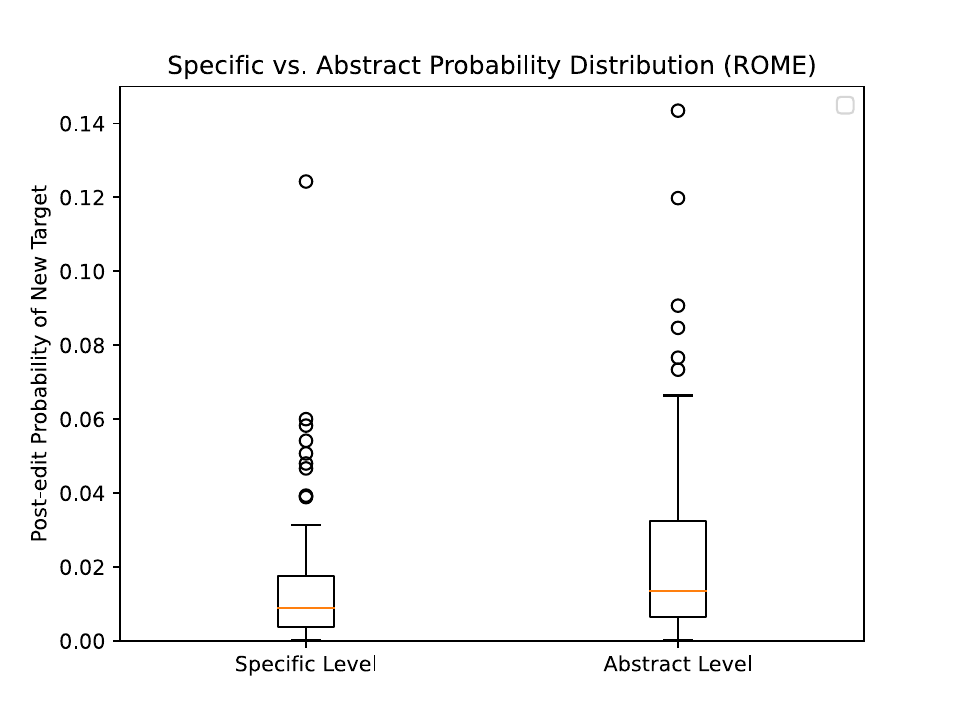}
\includegraphics[width=6cm]{gpt2xl_memit_h.pdf}
% \fbox{\rule[-.5cm]{0cm}{4cm} \rule[-.5cm]{4cm}{0cm}}
\end{center}
\caption{Specific vs. abstract probability distribution (\textsc{HierarchyData}) on GPT2-XL using a. FT (first) b. LoRA (second) c. ROME (third) d. MEMIT (fourth). }
\label{all_gpt2xl_h}
\end{figure}

\begin{figure}[h]
\begin{center}
% \framebox[4.0in]{$\;$}
\includegraphics[width=6cm]{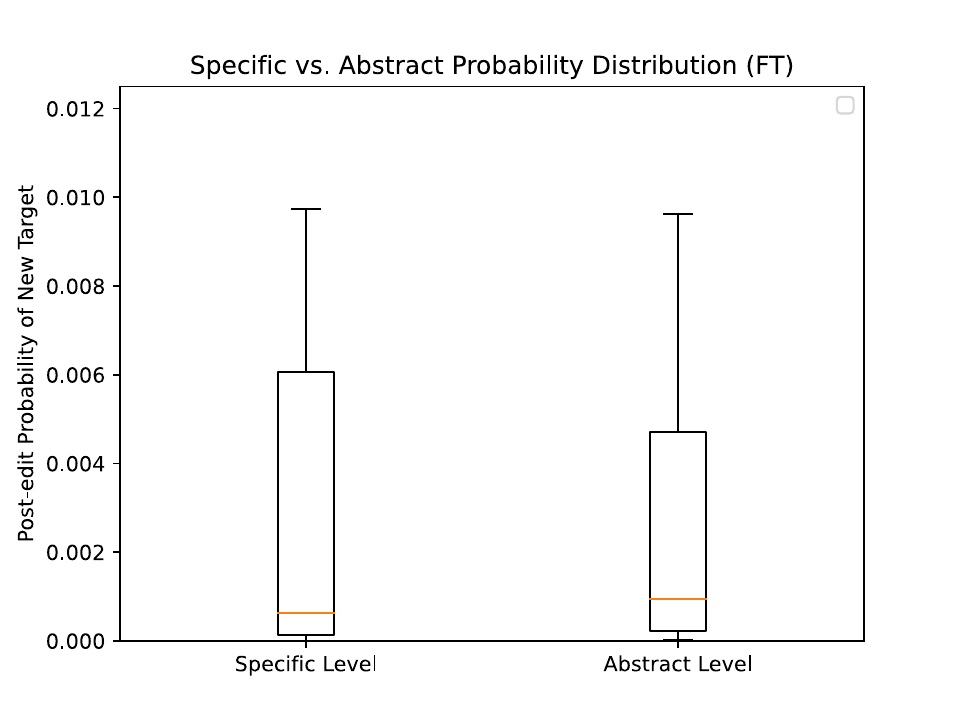}
\includegraphics[width=6cm]{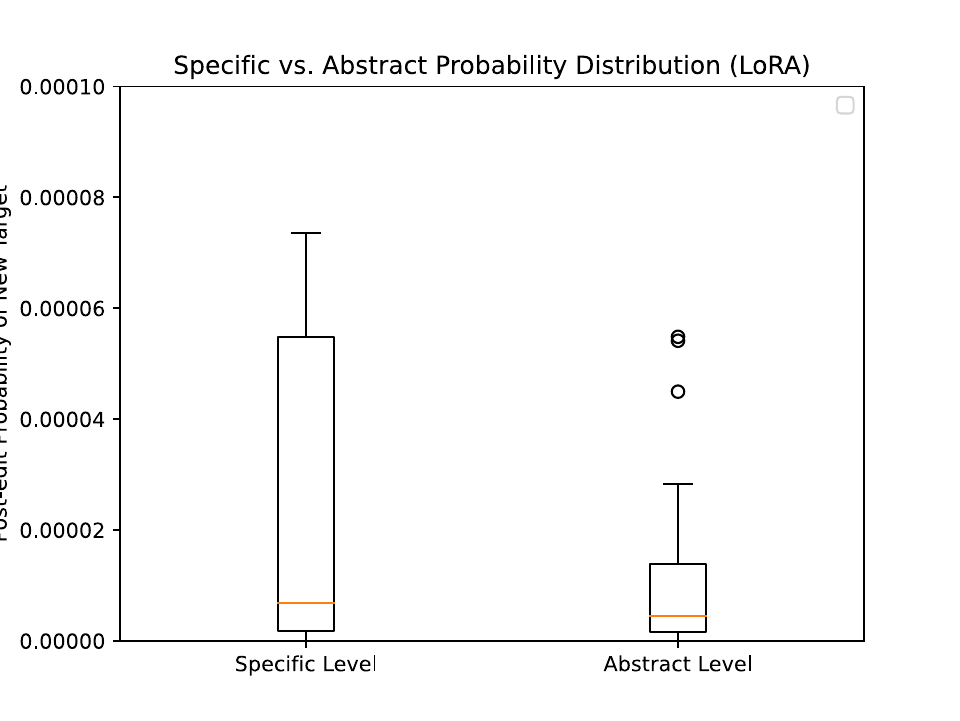}
\includegraphics[width=6cm]{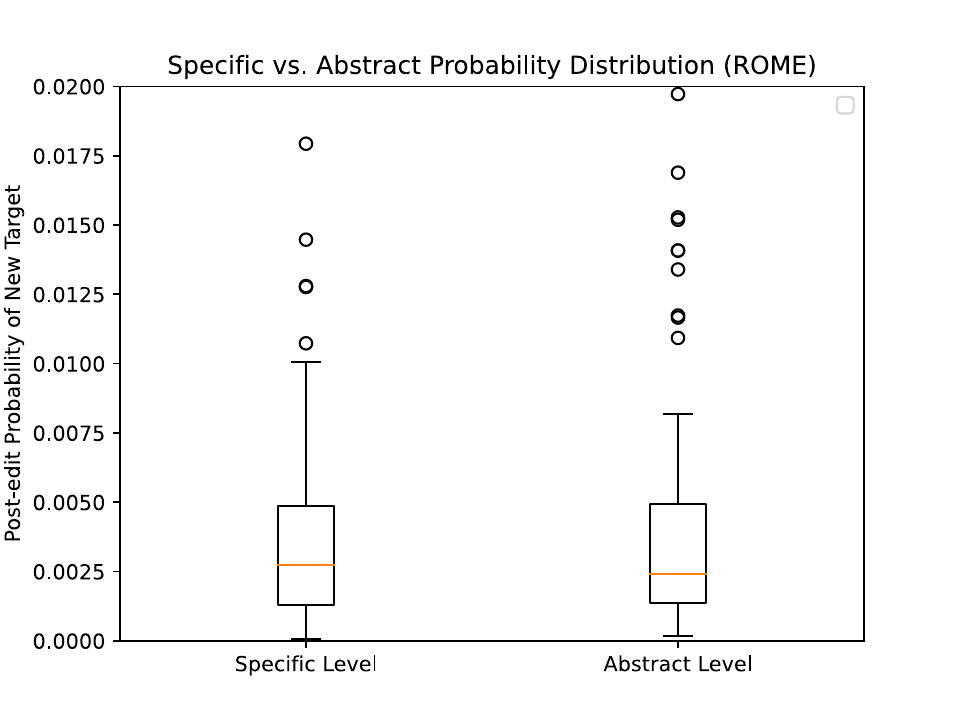}
\includegraphics[width=6cm]{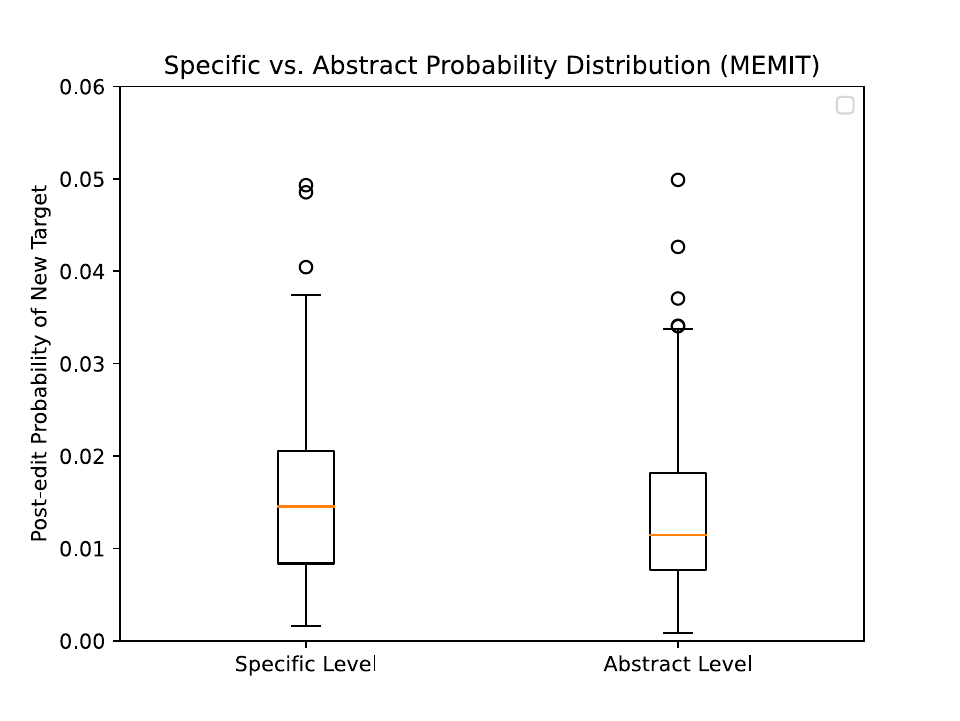}
% \fbox{\rule[-.5cm]{0cm}{4cm} \rule[-.5cm]{4cm}{0cm}}
\end{center}
\caption{Specific vs. abstract probability distribution (\textsc{HierarchyData}) on GPT-J(6B) using a. FT (first) b. LoRA (second) c. ROME (third) d. MEMIT (fourth). }
\label{all_gptj_h}
\end{figure}

We conduct a comparative analysis by plotting the efficacy distributions for data at both specific and abstract hierarchical levels, utilizing 198 groupings from the \textsc{HierarchyData} dataset—comprising an equal split of 99 specific-level instances and 99 abstract-level instances. Figure~\ref{all_gpt2l_h} showcases the box plot for editing methods applied to the GPT2-Large model. In a similar vein, Figure~\ref{all_gpt2xl_h} displays the box plot for techniques employed on the GPT2-XL model, while Figure~\ref{all_gptj_h} reveals the box plot corresponding to edits made on the GPT-J(6B) model.

\section{Pre-edit Specific vs. Abstract Probability Distribution in \textsc{HierarchyData}}
\label{sec:appendix:pre-edit-probability}

\begin{figure}[h]
\begin{center}
% \framebox[4.0in]{$\;$}
\includegraphics[width=6cm]{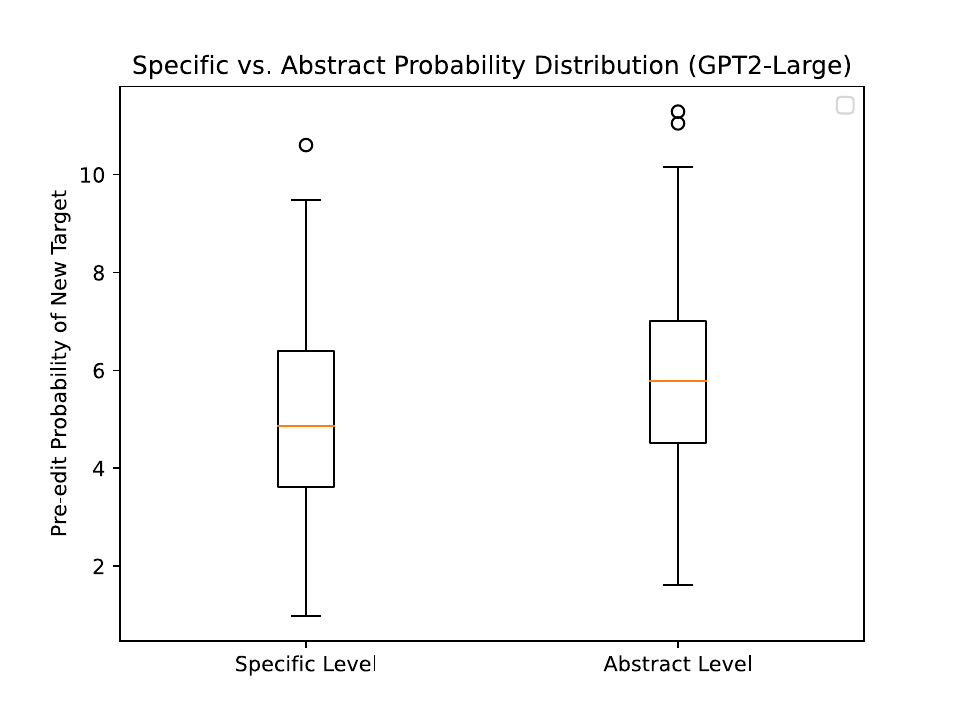}
\includegraphics[width=6cm]{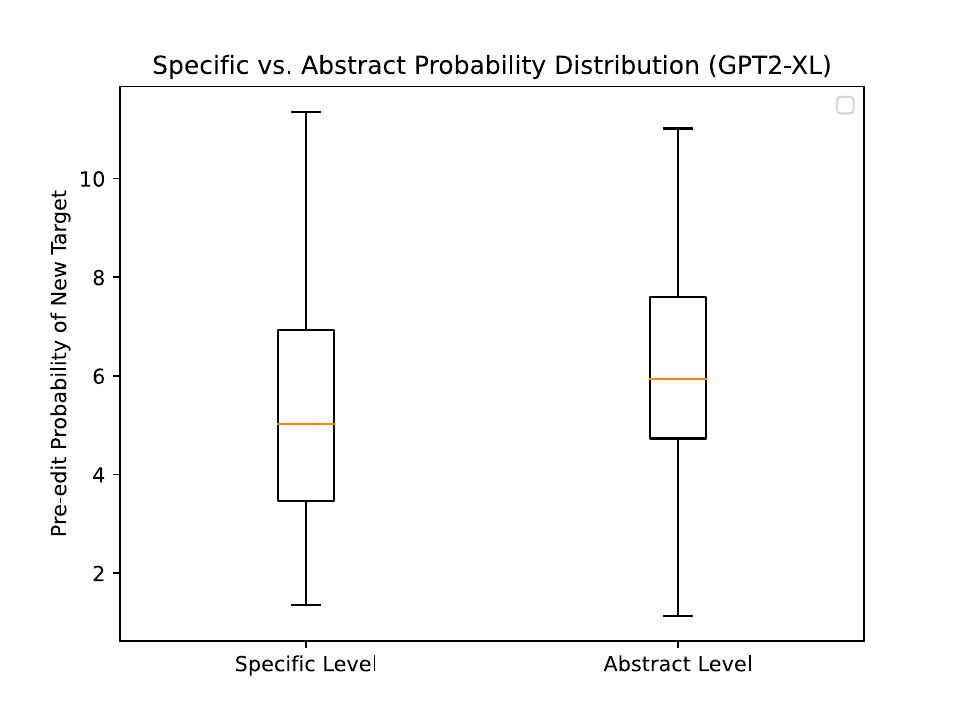}
\includegraphics[width=6cm]{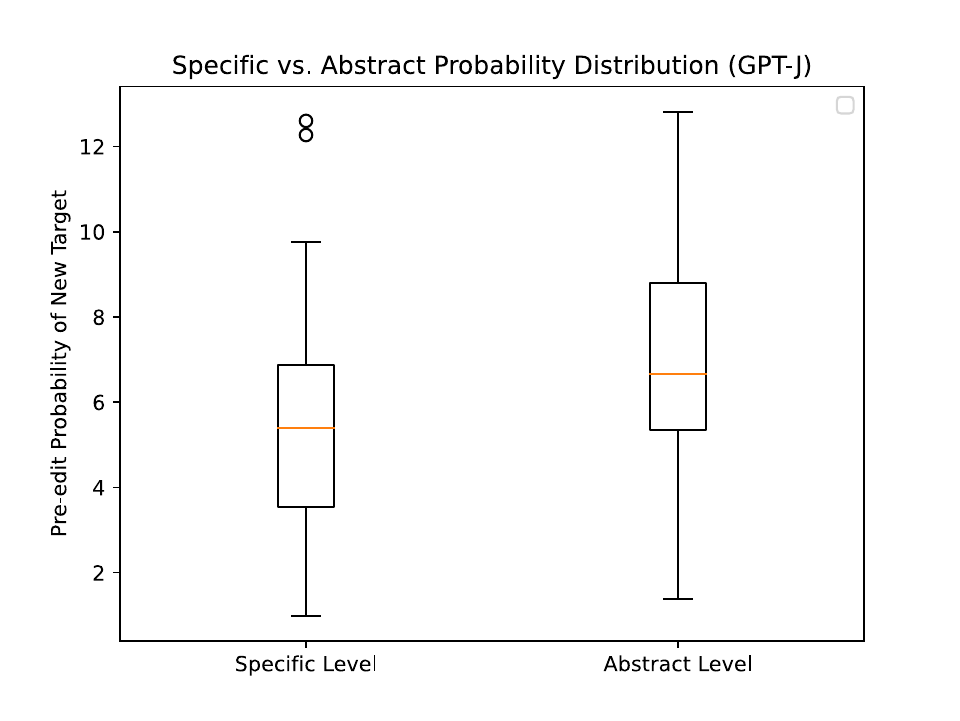}
% \fbox{\rule[-.5cm]{0cm}{4cm} \rule[-.5cm]{4cm}{0cm}}
\end{center}
\caption{Pre-edit specific vs. abstract probability distribution (\textsc{HierarchyData}) on a. GPT2-Large (first) b. GPT2-XL (second) c. GPT-J(6B) (third). }
\label{all_hp}
\end{figure}

We perform a comparative analysis of the perplexingness across both specific and abstract hierarchical levels by plotting their distributions. This analysis is based on 198 instances from the \textsc{HierarchyData} dataset, evenly divided between 99 specific-level and 99 abstract-level cases. Figure~\ref{all_hp} presents the box plots, illustrating the impact of editing methods on the GPT2-Large, GPT2-XL, and GPT-J(6B) models, thereby offering insights into the variation of perplexingness across different levels of hierarchy and models.

\end{document}